# Probabilistic Approach to Neural Networks Computation Based on Quantum Probability Model

## Probabilistic Principal Subspace Analysis Example

Marko V. Jankovic, Institute of Electrical Engineering “Nikola Tesla”, Belgrade, Serbia

***Abstract*—**In this paper, we introduce elements of probabilistic model that is suitable for modeling of learning algorithms in biologically plausible artificial neural networks framework. Model is based on two of the main concepts in quantum physics – a density matrix and the Born rule. As an example, we will show that proposed probabilistic interpretation is suitable for modeling of on-line learning algorithms for PSA, which are preferably realized by a parallel hardware based on very simple computational units. Proposed concept (model) can be used in the context of improving algorithm convergence speed, learning factor choice, or input signal scale robustness. We are going to see how the Born rule and the Hebbian learning rule are connected.

M. V. Jankovic is with the EE Institute “Nikola Tesla”, Koste Glavinica 8a, Belgrade, Serbia (phone: +381-11-3691-447; e-mail: elmarkoni@ ieent.org).

# Probabilistic Approach to Neural Networks Computation Based on Quantum Probability Model

## Probabilistic Principal Subspace Analysis Example

Marko V. Jankovic, Institute of Electrical Engineering "Nikola Tesla", Belgrade, Serbia

*Abstract*— In this paper, we introduce probabilistic interpretation of the Principal Subspace Analysis (PSA) based on two of the main concepts in quantum physics – a density matrix and the Born rule. We analyze the proposed probabilistic model in the artificial neural network framework. In other words, we are looking for suitable probabilistic interpretation of on-line learning algorithms for PSA, which are preferably realized by a parallel hardware based on very simple computational units. First, we give a probabilistic interpretation of the Modulated Hebb-Oja learning method. Later, we show how the proposed method can be generalized in order to obtain a number of PSA algorithms. At the end, we present some experimental results to illustrate the usefulness of the proposed probabilistic PSA model.

## I. INTRODUCTION

Principal subspace analysis (PSA) and principal component analysis (PCA) (e.g. [23]) are the well established techniques for dimensionality reduction, and chapters on the subject may be found in numerous texts on multivariate data analysis. Examples of its many applications include data compression, image processing, visualization, exploratory data analysis, pattern recognition, and time series prediction.

In literature, usually, we can find that definitions of PSA and PCA are not given separately. Here, we will give two most frequently cited definitions, and then we will give a short discussion:

**PCA** can be defined as the orthogonal projection of the data onto a lower dimensional linear space, known as the principal subspace, such that the variance of the projected data is maximized (Hotteling, 1933 [15]).

In the context of neural networks, this definition leads to the Subspace Learning Algorithm (SLA) [30] and its derivatives.

**Equivalently**, it can be defined as the linear projection that minimizes the average projection cost, defined as the mean square distance between the data points and their projections (Pearson, 1901 [31]).

In the context of neural networks, this definition represents the starting point of the Modulated Hebb-Oja algorithm [18] and its derivatives.

Here we will point out two things:

- In both **cases** definitions of PSA but not PCA are presented. Although PSA and PCA are similar techniques, they should be considered as different. First of all, the solution of PCA is represented by a discrete set (any permutation of eigenvectors of the input covariance matrix), while the solution of PSA is represented by a continuous set. Another important difference is that the input data covariance matrix is diagonal only in the coordinate system which is defined by eigenvectors themselves, but not in any other rotated system. Also, if the data belongs to the low-dimensional subspace of the input space, it will have a sparse representation if the PCA is applied, while this, generally, will not be achieved if the PSA is applied. Here, it has to be stressed that this can be achieved only if the input data is unimodal. If the input data is multimodal, then PCA will not result in sparse representation either.
- In the second definition, we used a term equivalently – here, we want to stress out that definitions are equivalent form the mathematical point of view. However, from the realization point of view, these two definitions are quite different (see [18]). The first one requires a big number of global calculation circuits, while the second one requires only one or two global calculation circuits.

PSA also has another convenient interpretation that is perhaps less well known. It is the probabilistic interpretation given in [34]. Although it is called probabilistic PCA, it represents probabilistic PSA model. In order to achieve different objectives, it is further generalized in [3], [7], [10], [12] or [35]. All these models are based on some distribution of latent variable model, so they can be used also in the case when we have problem of missing data. One of the most powerful algorithms for iterative PSA calculation is EM type algorithm proposed in [34, 35]. The EM algorithm can be modified in such a way as to yield principal components, sorted in descending order of the corresponding eigenvalues, directly [1], or PCA method. This model can be linked to probabilistic coupled generative model [1]. Proposed models resulted in great number of useful applications in the context of clustering, density modeling, outlier detection etc. Common characteristic of all mentioned probabilistic models is that they are useful for generation of off-line algorithms (they require storage of all data or their covariance matrix). So, they usually require powerful hardware for their realization. In the case that we need tracking of slow changes of correlations in the input data or in updating eigenvectors with new samples these algorithms are computationally expensive. Also, they rely on the assumption that input data is sampled from some distribution (Gaussian, exponential family, t-distribution, or similar) or their mixtures. So, it is not clear how these probabilistic models can be used in the context of artificial neural networks (or on-line learning algorithms) for PSA/PCA calculation, and/or, could they be realized on a very simple computational structures like linear neural networks are.

In this paper, we will proposed a sort of "quantum" probabilistic PSA model which relies on a very small number of assumptions and that is suitable, as we are going to see, for modeling the on-line type learning algorithms for calculation of PSA/PCA. We will illustrate the usefulness of the proposed concept (model) in the context of algorithm convergence speed, learning factor choice, or input signal scale robustness.

Why are we still interested in standard linear neural networks approach? Artificial neurons and neural networks have been shown to perform PCA when gradient ascent (descent) learning rules are used, which is related to the constrained maximization (minimization) of some objective functions [4, 9, 18, 29]. Due to their low complexity, such algorithms and their implementation in neural networks are potentially useful for the tracking of slow changes of correlations in the input data or in updating eigenvectors with new samples. They, also, could be successfully implemented on platforms like graphics processing unit (GPU). Furthermore, this kind of networks can be used for calculations of PCA and PSA for the higher dimensional systems than those solutions that require the storage of the whole covariance matrix. Another reason is that neural networks approach based on biologically plausible learning rules are still useful for the research in which the goal is to make computational models that emulate some of the brain circuitry.

In recent years, many Principal Subspace Analysis (PSA) algorithms have been proposed and studied in literature [4, 8, 13, 16, 18, 29]. Some of these PSA algorithms were modified in order to derive parallel PCA algorithms. Usually, the modification was performed by introduction of some asymmetry (inhomogenity) or nonlinearity in the original PSA algorithm that is not considered desirable from the point of view of the implementation of those algorithms in parallel hardware. Rare examples of a fully homogeneous algorithm are the bigradient algorithm proposed in [26], and the time-oriented hierarchical method (TOHM), proposed in [16, 19]. For a comprehensive review of the known parallel, as well as sequential PCA algorithms, see e.g. [9].

Recently, in [18], a new biologically inspired PSA method, named Modulated Hebb-Oja (MHO) learning rule, has been introduced. Major objectives for the method derivation were:

— to obtain a network which has a learning rule for individual synaptic efficacy that requires the least possible amount of explicit information about the other synaptic efficacies, especially those related to other neurons (in other words the locality of the calculation was considered very important);

— to minimize the neural hardware that is necessary for implementation of the proposed learning rule. Especially, the global calculation "neural circuitry" should be very simple, and the number of global calculation circuits should be as small as possible.

In this paper, we will give a probabilistic view of the PSA problem, starting with a probabilistic model of MHO learning algorithm. The probabilistic view is based on a new geometric interpretation of probability based on the Born rule [38]. Also, we will shortly introduce recently proposed probabilistic PCA model. In the case of on-line implementation, we will use constraint optimization on two time scales, as is done in the TOHM method, for instance. From the aspect of artificial neural networks, the choice of different realization concepts has a direct impact on the algorithm's convergence speed, preciseness, complexity of plausible hardware realization or biological plausibility. In an off-line context, probabilistic PCA can be seen as a successive optimization problem. In Section II, the Born rule will be introduced. Geometric interpretation of the Born rule will be given in Section III. The Born rule in the artificial neural networks framework is introduced in Section IV. In section V we introduce of new probabilistic PSA. In that section we will see what is the connection between the Born rule and the Hebbian learning rule. Experimental results are presented in Section VI. Section VII gives the conclusion remarks.

## II. Quantum Probability Model and Born Rule

In this section, we give a short introduction of the quantum probability model and the Born rule, based on a similar section in [38]. A slightly different approach can be found in [37]. In quantum mechanics, the transition from a deterministic description to a probabilistic one is done using a simple rule, termed the Born rule.

In quantum mechanics, the Born rule is usually taken as one of the axioms. However, this rule has well established foundations. Gleason's theorem [14] states that the Born rule is the only consistent probability distribution for a Hilbert space structure. Wooters [39] has shown that by using the Born rule as a probability rule, the natural Euclidean metrics on a Hilbert space coincides with a natural notion of a statistical distance. A short review for some other justifications of the Born rule can be seen in [38].

The quantum probability model takes place in a Hilbert space H of finite or infinite dimension. A state is represented by a positive semidefinite linear mapping (a matrix $\boldsymbol{\rho}$) from this space to itself, with a trace of 1, i.e. $\forall \boldsymbol{\Psi} \in H$ $\boldsymbol{\Psi}^T \boldsymbol{\rho} \boldsymbol{\Psi} \geq 0$, $\mathrm{Tr}(\boldsymbol{\rho}) = 1$. Such $\boldsymbol{\rho}$ is self adjoint and is called a density matrix.

Since $\boldsymbol{\rho}$ is self adjoint, its eigenvectors $\boldsymbol{\Phi}_i$ are orthonormal, and since it is positive semidefinite, its eigenvalues $p_i$ are real and nonnegative $p_i \geq 0$. The trace of a matrix is equal to the sum of its eigenvalues, therefore $\sum_i p_i = 1$.

The equation $\boldsymbol{\rho} = \sum_i p_i \boldsymbol{\Phi}_i \boldsymbol{\Phi}_i^T$ is interpreted as "the system is in state $\boldsymbol{\Phi}_i$ with probability $p_i$". The state $\boldsymbol{\rho}$ is called the pure state if $\exists i$ s.t. $p_i = 1$. In this case, $\boldsymbol{\rho} = \boldsymbol{\Psi}\boldsymbol{\Psi}^T$ for some normalized state vector $\boldsymbol{\Psi}$, and the system is said to be in state $\boldsymbol{\Psi}$. So, the most general density operator is in the form $\boldsymbol{\rho} = \sum_k p_k \boldsymbol{\Psi}_k \boldsymbol{\Psi}_k^T$ where the coefficients $p_k$ are nonnegative and add up to one, and $\boldsymbol{\Psi}_k$ represent pure states. We can see that this decomposition is not unique.

A measurement M with an outcome $z$ in some set $\boldsymbol{Z}$ is represented by a collection of positive definite matrices $\{\boldsymbol{m}_z\}_{z \in \boldsymbol{Z}}$ such that $\sum_{z \in \boldsymbol{Z}} \boldsymbol{m}_z = \mathbf{1}$($\mathbf{1}$ is being the identity matrix in H). Applying measurement M to state $\boldsymbol{\rho}$ produces the outcome $z$ with probability

$$p_z(\boldsymbol{\rho}) = \mathrm{trace}(\boldsymbol{\rho} \boldsymbol{m}_z) \qquad (1)$$

This is the Born rule. Most quantum models deal with a more restrictive type of measurement called the von Neumann measurement, which involves a set of projection operators $\boldsymbol{m}_a = \boldsymbol{a}\boldsymbol{a}^T$, for which $\boldsymbol{a}^T \boldsymbol{a}' = \delta_{\boldsymbol{aa}'}$.

In a modern language, von Neumann's measurement is a conditional expectation onto a maximal Abelian subalgebra of the algebra of all bounded operators acting on the given Hilbert space. As before, $\sum_{\boldsymbol{a}\in \boldsymbol{M}} \boldsymbol{a}\,\boldsymbol{a}^{\mathrm{T}} = \mathbf{1}$. For this type of measurement, the Born rule takes a simpler form: $p_{\boldsymbol{a}}(\boldsymbol{\rho})=\boldsymbol{a}^{\mathrm{T}}\boldsymbol{\rho}\boldsymbol{a}$. Assuming $\boldsymbol{\rho}$ is a pure state this can be simplified further to

$$p_{\boldsymbol{a}}(\boldsymbol{\rho}) = (\boldsymbol{a}^{\mathrm{T}}\boldsymbol{\Psi})^2. \qquad (2)$$

So, we can see that, if the state is $\boldsymbol{\rho}$, the probability of the outcome of the measurement will be $\boldsymbol{a}$, which is actually defined by the cosine square of the angle between vectors $\boldsymbol{a}$ and $\boldsymbol{\Psi}$, or $p_{\boldsymbol{a}}(\boldsymbol{\rho})=\cos^2(\boldsymbol{a},\boldsymbol{\Psi})$.

## III. JoyStick Probability Selector

In this section, we will give one novel, but simple interpretation of the probability that is related to the Born rule. Here, we will assume that we are dealing with a finite dimensional discrete variable.

For the moment, let's assume that we are dealing with a discrete two dimensional variable. We can associate it with a coin tossing. Assume further that two possible outcomes of our experiment are represented by the dummy variables {01} and {10}. If we represent our coin as a unit norm vector in the two dimensional space (we will call that vector JoyStick Probability Selector or JSPS), then we can have the following simple geometric interpretation given in Fig. 1.

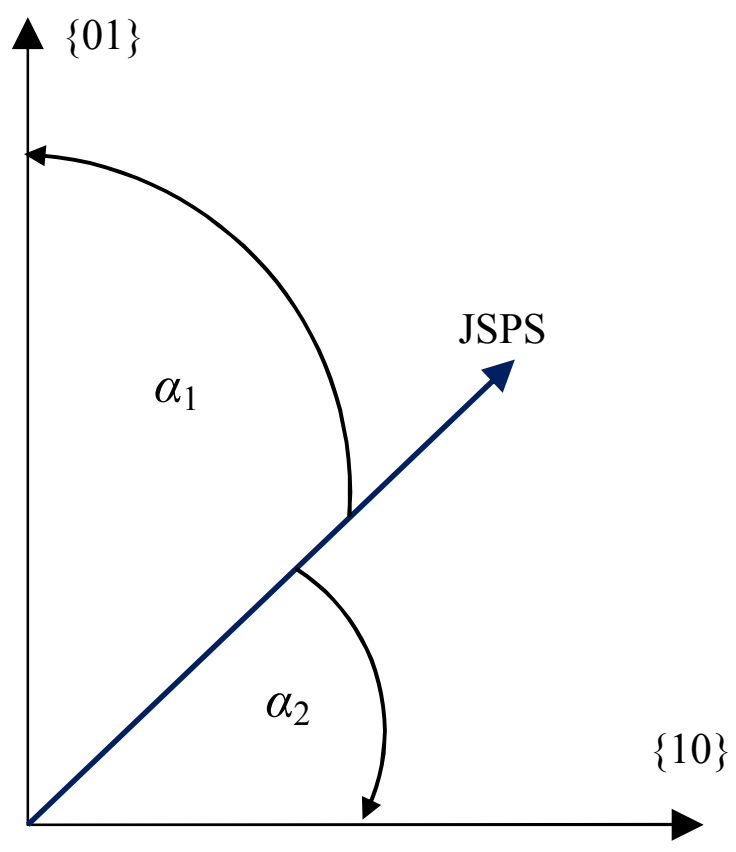

Fig 1. JoyStick Probability Selector

Now, we will suggest that the probability of the outcome {01} is equal to the cosine square of angle $\alpha_1$, while the probability of outcome {10} is equal to the cosine square of angle $\alpha_2$. It is not difficult to see that $\cos^2(\alpha_1) + \cos^2(\alpha_2) = 1$. We can see that the probability of the particular outcome of the experiment (in this case the coin toss) is equal to the inner product of the unit norm JSPS and the unit norm vector which represents that outcome. Then, we can see that this coincides with the Born rule interpretation for the case of a pure state and the von Neuman measurement system. When the state (JSPS) vector collapses to one of the states, it is not possible to give information of the other state, which is consistent with some quantum mechanics results.

We can check what will happen if our discrete variable is of the dimension 3. In that case, our system can be represented as Fig. 2. Now, we have the three possible outcomes of the experiment that are represented by dummy variables {001}, {010} and {100}. Again, we have the JSPS vector which represents the status of our variable before we perform the measurement. Again, the probability of the outcome is given by the cosine square of the angle between JSPS and the particular outcome vector.

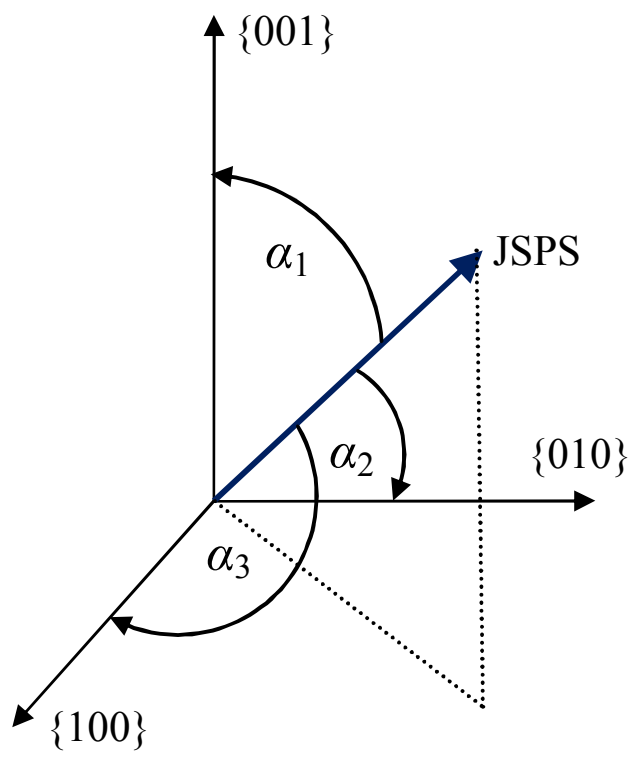

Fig. 2 A 3-D example

It is not difficult to check that

$$\sum_{i=1}^{3} \cos^2(\alpha_i) = 1,, \qquad (3)$$

which follows from generalized Pythagorean Theorem, or Parseval's Theorem. For any 3-D vector whose norm is *r* we have

$$\sum_{i=1}^{3} r^2 \cos^2(\alpha_i) = r^2. \qquad (4)$$

This way of reasoning can be extended to any finite dimension *D*. It can be extended, under some assumptions, to infinite dimensional cases (see e.g. [24] and [25]), but it will not be discussed here in detail.

From JSPS interpretation of the Born rule, we can see that every vector in finite dimensional Hilbert space represents a discrete distribution which is defined by unit norm vector in the same direction. Also, any discrete probability distribution can be represented as a vector in a Hilbert space of proper dimension. The norm of the vector can be interpreted in several ways (we will not discuss it here in details). An interpretation, that is going to be used in the text that follows, is that norm of the vector is related to the probability of the appearance of that vector.

## IV. BORN RULE IN NEURAL NETWORK CONTEXT

In this section, we will give some explanations and definitions that will be used in the following sections. It will be explained here, how some probabilistic concepts can be used in a neural network framework, or how neural networks can be used in a probabilistic framework that is basically based on the Born rule.

The basic single layer feedforward artificial neural network is depicted in Fig. 3. The output of the *n*-th output unit $y_n$ ($n$=1, 2,…, $N$) of a layer of parallel linear artificial neurons is given as

$$y_n(i) = \boldsymbol{w}_n(i)^{\mathrm{T}}\, \boldsymbol{x}(i),$$

with $\boldsymbol{x}(i)$ denoting a $K$-dimensional zero-mean input vector of the network and $\boldsymbol{w}_n(i)$ denoting a weight vector of the $n$-th output unit, and $i$ represents sampling instances $iT$, where $T$ is a sampling period. The output vector $\boldsymbol{y}$ is defined as

$$\boldsymbol{y}(i) = \boldsymbol{W}(i)^{\mathrm{T}}\, \boldsymbol{x}(i). \quad (5)$$

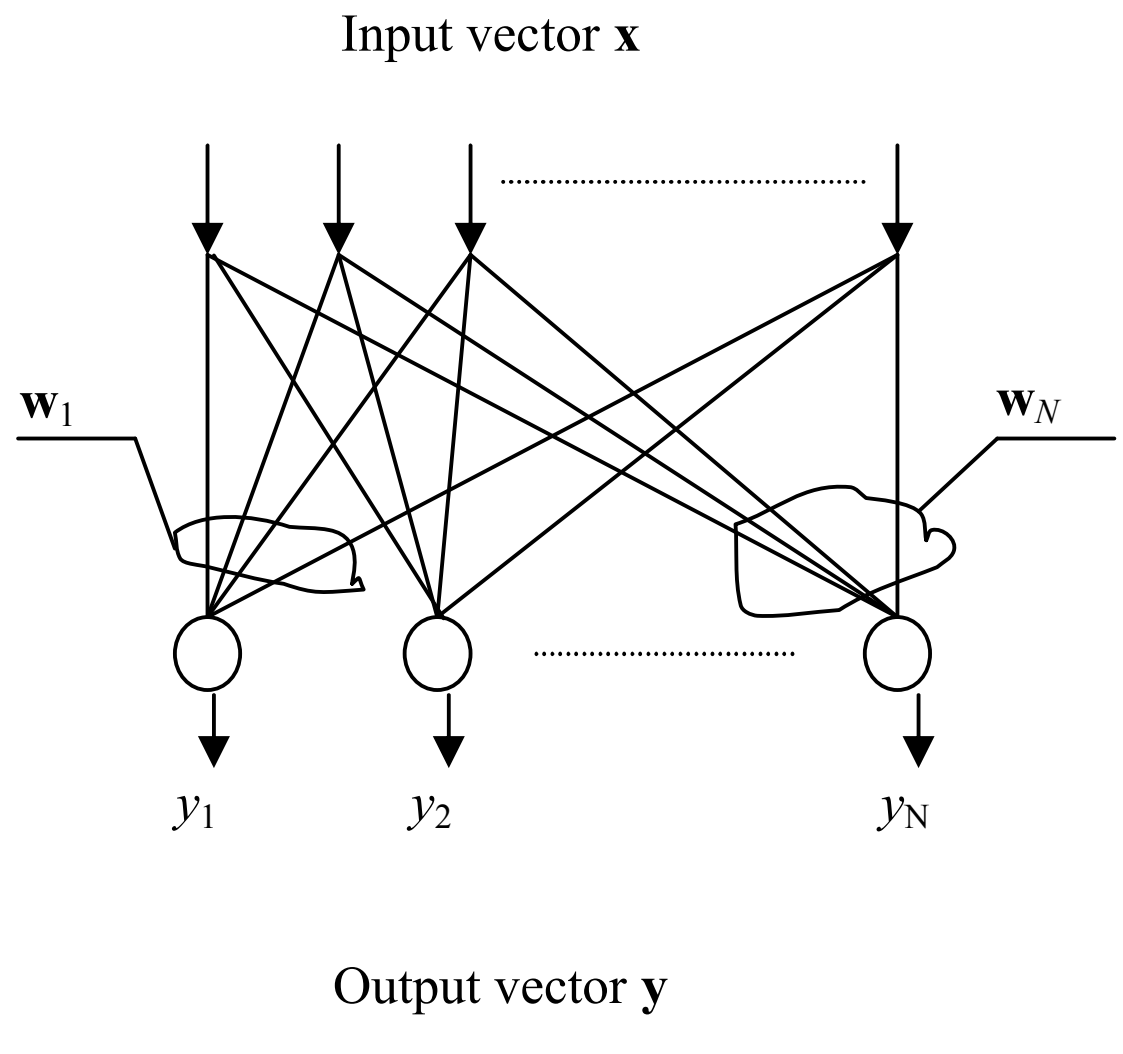

Fig. 3. A layer of parallel linear artificial neurons

In the usual interpretation, based on specific requirements, e.g. minimization of some cost function, matrix $\boldsymbol{W}$ is changed (trained) in the process of learning, according to some adopted learning rule.

Here we will give a slightly different interpretation. Again, we will consider a Hilbert space H of a finite dimension. "State vectors" are defined by the input data vector $\boldsymbol{x}_k$ and we can imagine that every vector $\boldsymbol{x}_k$ is available in a big enough number of copies (clones), so we can perform as many simultaneous measurements as we want. A measurement M with an outcome $\boldsymbol{w}_n$ in some set $\boldsymbol{W}$ is represented by a collection of positive definite matrices $\{\boldsymbol{m}_{w_n}\}_{w_n \in W}$ such that $\boldsymbol{m}_{w_n}=\boldsymbol{w}_n\boldsymbol{w}_n^{\mathbf{T}}$, so $\sum_{w_n \in W} =\boldsymbol{W}\boldsymbol{W}^{\mathbf{T}}$, which is not necessarily equal to the identity matrix on H. This means that the sum of the probabilities of the particular outcomes does not have to be equal to one – in other words, sometimes we will work with improper discrete distributions. Also, measures like entropy and divergence will be applied to improper probability distributions, or to a mixture of proper and improper probability distributions. In the following sections, we will point out that in the adopted framework, this will not affect the final result. Applying measurement M to state $\boldsymbol{x}_k$ produces outcome $\boldsymbol{w}_n$ with the probability (the Born rule)

$$p(\boldsymbol{w}_n \mid \boldsymbol{x}_k) \overset{def}{=} \cos(\boldsymbol{w}_n, \boldsymbol{x}_k),$$

regardless of the norm of the vectors $\boldsymbol{w}_n$ and $\boldsymbol{x}_k$. In the following text, we will consider only vectors $\boldsymbol{w}_n$ that have unit norms. This means

$$p(\boldsymbol{w}_n \mid \boldsymbol{x}_k) = \frac{(\boldsymbol{w}_n^{\mathrm{T}} \boldsymbol{x}_k)^2}{\|\boldsymbol{x}_k\|^2} \quad .$$

Also, if we apply $N$ simultaneous measurements $\mathrm{M}^N$ to the state $\boldsymbol{x}_k$ we obtain outcome $\boldsymbol{W}$ with the probability

$$p(\boldsymbol{W} \mid \boldsymbol{x}_k) \overset{def}{=} \sum_{n=1}^{N} p(\boldsymbol{w}_n \mid \boldsymbol{x}_k).$$

Here, it is assumed that the outcome of each measurement is different. We define the joint probability of the state $\boldsymbol{x}_k$ and outcome $\boldsymbol{W}$ obtained by simultaneous multiple measurement $\mathrm{M}^{\mathrm{N}}$ on state $\boldsymbol{x}_k$, $p(\boldsymbol{W}, \boldsymbol{x}_k)$ as

$$p(\boldsymbol{W}, \boldsymbol{x}_k) \overset{def}{=} p(\boldsymbol{W} \mid \boldsymbol{x}_k) p(\boldsymbol{x}_k) . (6)$$

Now, without loss of generality, let's assume that we are dealing with a random variable $\boldsymbol{x}$ that takes realizations from a set of observed $K$-dimensional zero-mean data vectors $\{\boldsymbol{x}_k\}$, $k \in \{1, \ldots, N_{\mathrm{sample}}\}$, which are sampled from some distribution in time instants $t = kT$ where $k$ is already defined and $T$ represents the sampling period. Then, we can define $p(\boldsymbol{x}{=}\boldsymbol{x}_k \mid t{=}kT)$ as

$$p(\boldsymbol{x}_k) \overset{def}{=} \frac{\|\boldsymbol{x}_k\|^2}{\sum_{i=1}^{N_{sample}} \|\boldsymbol{x}_i\|^2}, \quad (7)$$

where $N_{sample}$ represents the overall number of samples that are going to be analyzed. It is interesting to note that the only thing that we can conclude about the $p(\boldsymbol{x}_k)$ is that it is proportional to $\|\boldsymbol{x}_k\|^2$. The sum in the denominator represents the energy of samples that are going to be analyzed – we actually do not know the value of that sum in any, but the final moment. However, we know that it represents some constant. We can easily see that the adopted probability measure fulfils the two conditions that are required for the probability function $f(\boldsymbol{z})$ (in our case $p(\boldsymbol{z})$) to be considered as a modified generalized probability measure [37]:

1. For each $\boldsymbol{z}$, $0 \leq f(\boldsymbol{z}) \leq 1$ ,
2. $\sum_{\mathrm{i}} f(\boldsymbol{z}_i) = 1$.

In this definition, orthonormallity is not explicitly required in order that the coefficients $f(\boldsymbol{z}_i)$ sum up to one. However, from the JSPS introduction, we can see that it is always implicitly present. Also, we know from the definition of density operator, which represents a mixture of the pure states (see Section II), that the trace of the density matrix has to be 1. Here, we will consider all vectors as "oriented energies" or

$$\boldsymbol{x}_k = \|\boldsymbol{x}_k\| \frac{\boldsymbol{x}_k}{\|\boldsymbol{x}_k\|} = \|\boldsymbol{x}_k\| \boldsymbol{xn}_k ,$$

where the norm of the vector $\|\boldsymbol{x}_k\|$, represents the square root of the energy contained in the vector $\boldsymbol{x}_k$, and the orientation represents some unit norm vector $\boldsymbol{xn}_k$, which represents some pure state. In that case, we can see that the statistical description of our system is represented by the density matrix $\boldsymbol{\rho}$

$$\boldsymbol{\rho} = \sum_k p_k \boldsymbol{xn}_k \boldsymbol{xn}_k^{\mathrm{T}} \ ,$$

as a statistical mixture of pure states $\boldsymbol{xn}_k$, and $p_k=p(\boldsymbol{x}_k)$ are defined by (7). We can see that

$$\boldsymbol{\rho} = \frac{N_{sample}}{\sum_{i=1}^{N_{sample}} \|\boldsymbol{x}_i\|^2} \boldsymbol{C} \ ,$$

where $\boldsymbol{C}$ is input signal covariance matrix. Obviously, the matrix $\boldsymbol{\rho}$ and the matrix $\boldsymbol{C}$ have the same eigenvectors. In some sense, proposed approach to input sample probability modeling, is similar to the Dirichlet process model, which is usually used to model a stream of symbols where the vocabulary of symbols is not limited [27]. In the Dirichlet process case, we have one parameter $\alpha$ which is constant. In our case, it is not constant but it is defined by the energy of the new incoming symbol (sample). Also, in our case every incoming symbol is treated as a new one – so, initially we do not consider that the same samples in different moments are the same (e.g. we can make analogy to spin of a particle). That means that our input space is initially considered as an augmented space of dimension $N_{sample}$. When we observe all samples, we can switch to the "deformed" space of dimension $N_d$, which is defined by the number of different energy levels $d$, or some other "deformed" space which we want to analyze. If we want to calculate "marginal" probability $p(\boldsymbol{x}=\boldsymbol{x}_s)$ (i.e. probability of observing a certain value $\boldsymbol{x}_s$ no matter what the time instant $i$ is) we can do it as

$$p(\boldsymbol{x}=\boldsymbol{x}_s)=\sum_{i=1}^{N_{sample}}\delta(\boldsymbol{x}_i,\boldsymbol{x}_s)p(\boldsymbol{x}_i), \qquad (8)$$

where $\delta$ represents the Kronecker $\delta$ function.

In the proposed context, the learning algorithm applied to the neural network has a basic task to find the measurement system in which input data is "best explained", or have the features that are specified. As an example, principal component analysis will search for the measurement (or we can say coordinate) system in which the input data covariance matrix is diagonal. Or, using the quantum mechanics terms, we are trying to find projective measurement which does not increase the von Neumann entropy of our density matrix [28]. In that case, data that belongs to some low dimensional subspace have sparse representation.

## V. Probabilistic PSA

Now, we will give a new definition of probabilistic PSA (having in mind the definitions given in the previous section):

**Definition 1:** PSA can be defined as a problem of minimization of some symmetric divergence/distance between the probability distribution of the input signal $p(\boldsymbol{x})$ and the joint probability distribution $p(\boldsymbol{W}, \boldsymbol{x})$ of the input signal $\boldsymbol{x}$ and the outcome $\boldsymbol{W}$, obtained by the simultaneous multiple measurement $\mathrm{M}^N$ on state (input signal) $\boldsymbol{x}$.

In the next subsection we are going to explain one possible solution of the PSA problem, starting from Definition 1 and adopting a particular symmetric divergence measure.

*A. Probabilistic PSA*

Without loss of generality, let's assume that we are dealing with a set of observed $K$-dimensional zero-mean data vectors $\{\boldsymbol{x}_i\}$, $i \in \{1, \ldots, N_{\text{sample}}\}$ which are sampled from some distribution in time instants $t = iT$, where $i$ is already defined and $T$ represents the sampling period. We will denote $\boldsymbol{w}_1, \boldsymbol{w}_2, \ldots, \boldsymbol{w}_N$ the **unit norm** column vectors of matrix $\boldsymbol{W}$ ($N \leq K$). Now, we are going to define probabilistic PSA. In order to do that, we are going to define the symmetric distance, which we will call quadratic variational divergence

$$\mathrm{D}_{\mathrm{qV}}\left(p(\boldsymbol{x}) \,\|\, p(\boldsymbol{W}, \boldsymbol{x})\right) = \mathrm{E}\left(\left(p(\boldsymbol{x}) - p(\boldsymbol{W}, \boldsymbol{x})\right)^2\right) = \sum_i \left(p(\boldsymbol{x}_i) - p(\boldsymbol{W}, \boldsymbol{x}_i)\right)^2, \quad (9)$$

where E represents the empirical expectation operator, and $\boldsymbol{W}$ is a matrix whose columns have unit norms. This formulation represents only a class of algorithms that **do not** require the explicit orthonormality of matrix $\boldsymbol{W}$. Explicit orthonormality is sometimes not a desirable constraint, e.g. in biologically plausible neural networks, since the explicit orthonormality usually requires global computation. Matrix $\boldsymbol{W}$ with unit norm columns can be obtained by local calculations.

We can see that (9) can be written as

$$\begin{aligned}\mathrm{JS1}(\boldsymbol{W}) &= \mathrm{D}_{\mathrm{qV}}\left(p(\boldsymbol{x}) \| p(\boldsymbol{W},\boldsymbol{x})\right) = \mathrm{E}\left((p(\boldsymbol{x}) - p(\boldsymbol{W},\boldsymbol{x}))^2\right) \\ &= \frac{1}{\left(\sum_{j=1}^{N_{sample}} \|\boldsymbol{x}_j\|^2\right)^2}\left(\sum_{k=1}^{N_{sample}}\left(\boldsymbol{x}_k^T\boldsymbol{x}_k - \sum_{n=1}^{N}\boldsymbol{w}_n^T\boldsymbol{x}_k\boldsymbol{x}_k^T\boldsymbol{w}_n\right)^2\right), \quad (10)\end{aligned}$$

Now, we define the probabilistic PSA as a constrained minimization problem

$$\min_{\boldsymbol{W}}\left(\mathrm{JS1}(\boldsymbol{W})\right), \qquad (11)$$

under the constraint that $\boldsymbol{W}$ consists of a unit norm columns. So, we are looking for $\boldsymbol{W}$ which minimizes divergence $D_{qV}$. As it is shown later, the solution will be obtained for matrix $\boldsymbol{W}=\boldsymbol{W}^{\mathrm{JS1}}$, which represents the rotated matrix of matrix $\boldsymbol{U}$**,** which consists of eigenvectors $\boldsymbol{u}_i$ as columns, where $\boldsymbol{u}_i$ represent $N$ dominant eigenvectors of the covariance matrix $\boldsymbol{C} = \mathrm{E}(\boldsymbol{x}\boldsymbol{x}^{\mathrm{T}})$. We can drop the constant that represents the square of overall energy of the input samples, so, the minimization of JS1 is equivalent to the minimization of the following function JS1m

$$\mathrm{JS1m}(\boldsymbol{W}) = \sum_{k=1}^{N_{sample}}\left(\boldsymbol{x}_k^{\mathrm{T}}\boldsymbol{x}_k - \sum_{n=1}^{N}\boldsymbol{w}_n^{\mathrm{T}}\boldsymbol{x}_k\boldsymbol{x}_k^{\mathrm{T}}\boldsymbol{w}_n\right)^2 = \sum_{k=1}^{N_{sample}}\left(\boldsymbol{x}_k^{\mathrm{T}}\boldsymbol{x}_k - \boldsymbol{y}_k^{\mathrm{T}}\boldsymbol{y}_k\right)^2$$

where $\boldsymbol{y}_k$ is defined by (5). Also, from (9) and (10) we can see that

$$p(\boldsymbol{x}_k) \propto \boldsymbol{x}_k^{\mathrm{T}} \boldsymbol{x}_k \text{ and } p(\boldsymbol{W}, \boldsymbol{x}_k) \propto \boldsymbol{y}_k^{\mathrm{T}} \boldsymbol{y}_k. \quad (12)$$

In the following text, we will use this fact, so we can make equations simpler. It was shown in [18] that the minimization of function JS1m by a gradient descent method (in the single layer neural network framework) will yield the principal subspace of the input covariance matrix. In [18] that algorithm was named the MH(O) algorithm and can be represented by the following learning rule (δ represents the Kronecker δ function):

$$\boldsymbol{W}(i+1) = \boldsymbol{W}(i) + \gamma(i)\left(\boldsymbol{x}(i)^{\mathrm{T}} \boldsymbol{x}(i) - \boldsymbol{y}(i)^{\mathrm{T}} \boldsymbol{y}(i)\right)\left(\boldsymbol{x}(i)\boldsymbol{y}(i)^{\mathrm{T}} - (1 - \delta(K, N))\boldsymbol{W}(i)\mathrm{diag}\left(\boldsymbol{y}(i)\boldsymbol{y}(i)^{\mathrm{T}}\right)\right).$$

We can use some other symmetric divergence/distance measures and create different types of algorithms. For instance we can use symmetric f-divergences. F-divergences were introduced by Csiszar [11] and Ali and Silvey [2]. Also, some other divergences/distances (or their linear combination) can be used. Some examples are given in the following Table I.

Table I

| Name | Equation |
| --- | --- |
| Variational distance | $\mathrm{D_V}(P \parallel Q) = \mathrm{E}(\lvert p-q \rvert)$ |
| Quadratic variational divergence | $\mathrm{D_{qV}}(P \parallel Q) = \mathrm{E}((p-q)^2)$ |
| Jefrrey's J-divergence | $\mathrm{D_J}(P \parallel Q) = \mathrm{E}\left((p-q)\log\left(\frac{p}{q}\right)\right)$ |
| Jensen-Shannon divergence | $\mathrm{D_{JS}}(P \parallel Q) = \frac{1}{2}\mathrm{E}\left(p\log\left(\frac{2p}{p+q}\right) + q\log\left(\frac{2q}{p+q}\right)\right)$ |
| Hellinger distance | $\mathrm{D_H}(P \parallel Q) = \frac{1}{2}\mathrm{E}\left(\left(\sqrt{p}-\sqrt{q}\right)^2\right)$ |

<u>*Note 1*</u>: A different class of probabilistic PSA algorithms can be obtained by the maximization of JS1_d (***W***) defined in the following equation:

$$\max_{\boldsymbol{W}} \mathrm{JS1_d}(\boldsymbol{W}) = \max_{\boldsymbol{W}} \mathrm{E}(p(\boldsymbol{W}, x_k))$$

under the constraint that ***W*** is orthonormal matrix. In that case the PSA problem is very similar to the clustering problem (see e.g. [21], [38]). This approach was analyzed in [22]. The Subspace Learning Algorithm (SLA) [30], can be seen as a particular realization of that approach.

*Note 2*: Since the definition of $p(\boldsymbol{w}_n|\boldsymbol{x}_k)$ is based on the Born rule, we can see that any algorithm that is based on gradient minimization/maximization of the function which contains terms $p(\boldsymbol{w}_n|\boldsymbol{x}_k)$ will result in the learning rule which contains the Hebbian learning rule part.

*B. Probabilistic PCA*

In this section, we are going to give a definition of the probabilistic PCA that can be used for creation of symmetric PCA algorithms. There is also another possibility to create asymmetric algorithms as done in [20].

**Definition 2:** PCA can be defined as a problem of minimization of the entropy of the joint probability distribution $p(\boldsymbol{W}, \boldsymbol{x}_k)$ of the input signal $\boldsymbol{x}$ and outcome $\boldsymbol{W}$, obtained by the simultaneous multiple measurement $\mathrm{M}^N$ on state (input signal) $\boldsymbol{x}$, under the constraint that the matrix $\boldsymbol{W}$ spans the principal subspace of the input signal covariance matrix $\boldsymbol{C} = \mathrm{E}(\boldsymbol{x}\boldsymbol{x}^{\mathrm{T}})$. (Definition can be also based on terms like projective measurement and density matrix of input signal, but here we opted for simpler one.)

We can see that the proposed probabilistic definition of PCA, probabilistic PSA represents an integral part and it directly influences the characteristics of probabilistic PCA realizations, like algorithm convergence speed, locality of calculations, preciseness and so on. This probabilistic approach to PCA was recently proposed and analyzed in [22]. Here we will make only a brief review.

*Theorem:* (**Projective measurement cannot decrease entropy [28]**)

*Suppose $P_k$ is a complete set of orthogonal projectors and* $\boldsymbol{\rho}$ *is a density operator. Then the von Neumann entropy of the state* $\boldsymbol{\rho}' = \sum_k \boldsymbol{P}_k \boldsymbol{\rho} \boldsymbol{P}_k$ *of the system after the measurement is at least as great as the original entropy,*

$$S(\boldsymbol{\rho}') \geq S(\boldsymbol{\rho})$$

*with equality if the* $\boldsymbol{\rho} = \boldsymbol{\rho}'$ (which s only the case when projections matrices are defined by eigenvectors of $\boldsymbol{\rho}$).

*Von Neumann entropy is defined as:* $S(\boldsymbol{\rho}) = -\mathrm{tr}(\boldsymbol{\rho} \log \boldsymbol{\rho}) = -\Sigma_i \lambda_i \log(\lambda_i)$, where $(\lambda_i)$ represent eigenvalues of matrix $\boldsymbol{\rho}$.

Let's assume that the matrix $\boldsymbol{W}=\boldsymbol{W}^{\mathrm{JS1}}$ is an orthonormal matrix and that it spans the principal subspace. Then, we are going to analyze the following cost function

$$\mathrm{JS2}\left(\boldsymbol{W}^{\mathrm{JS1}}\right) = \mathrm{E}\left(S\left(p\left(\boldsymbol{W}^{\mathrm{JS1}}, \boldsymbol{x}_k\right)\right)\right), \qquad (13)$$

where S represents the "principal subspace" entropy (see the next paragraph). For instance, we can chose the Shannon entropy $S=S_S$, (although we chose some other definition of entropy, like the Tsallis [36] entropy, or their linear combination) so our cost function is expressed as

$$\mathrm{JS2}\left(\boldsymbol{W}^{\mathrm{JS1}}\right)=\mathrm{E}\left(\sum_{n=1}^{N} p\left(\boldsymbol{w}_n^{\mathrm{JS1}}, \boldsymbol{x}_k\right)\log\left(\frac{1}{p\left(\boldsymbol{w}_n^{\mathrm{JS1}}, \boldsymbol{x}_k\right)}\right)\right) \quad (14)$$

Now, the probabilistic PCA can be defined as the following constrained minimization problem

$$\min_{\boldsymbol{W}^{\mathrm{JS1}}} \mathrm{JS2}\left(\boldsymbol{W}^{\mathrm{JS1}}\right), \qquad (15)$$

under constraint that $\boldsymbol{W}^{\mathrm{JS1}}$ is orthonormal and spans the principal subspace of matrix $\boldsymbol{C}$. It can be shown that the solution of this optimization problem will be $\boldsymbol{W}^{\mathrm{JS1}}= \boldsymbol{UP}$, where, as before, $\boldsymbol{U}$ consists of $K$ principal eigenvectors $\boldsymbol{u}_i$ as columns, where $\boldsymbol{u}_i$ represent principal eigenvectors of the covariance matrix $\boldsymbol{C} = \mathrm{E}(\boldsymbol{xx}^{\mathrm{T}})$, and $\boldsymbol{P}$ is a permutation matrix. The proof is straightforward, and it is based on the Section III and the fact that projective measurement can increase but never decrease von Neumann entropy of a mixture state (see section 11.3 in [28], particularly, the equations (11.57) and (11.66)).

We will have to emphasize that, strictly speaking, equation (14) is correct only in the case $N=K$, since in the case $N < K$, $\sum_n p(\boldsymbol{w}_n^{\mathrm{JS1}}, \boldsymbol{x}_k) < 1$. In other words, in the case $N < K$, $p(\boldsymbol{w}_n^{\mathrm{JS1}}, \boldsymbol{x}_k)$ does not represent a proper probability function since it is not normalized. In the case $N < K$, we will define the term "principal subspace" entropy, $\mathrm{S}^{\mathrm{PS}}$. Here, we will analyze only the Shannon entropy case, but something similar can be done for the Tsallis or other types of entropy. We define

$$\mathrm{S}^{\mathrm{PS}}\left(\boldsymbol{W}^{\mathrm{JS1}}\right)=\sum_{n=1}^{N} p\left(\boldsymbol{w}_n^{\mathrm{JS1}}, \boldsymbol{x}_k\right)\log\left(\frac{1}{p\left(\boldsymbol{w}_n^{\mathrm{JS1}}, \boldsymbol{x}_k\right)}\right). \quad (16)$$

The "correct" definition of the Shannon entropy in the principal subspace can be given as

$$\mathrm{S}_{\mathrm{S}}\left(\boldsymbol{W}^{\mathrm{JS1}}\right)=\sum_{n=1}^{N} p^{\mathrm{PS}}\left(\boldsymbol{w}_n^{\mathrm{JS1}}, \boldsymbol{x}_k\right)\log\left(\frac{1}{p^{\mathrm{PS}}\left(\boldsymbol{w}_n^{\mathrm{JS1}}, \boldsymbol{x}_k\right)}\right) \quad (17)$$

where

$$p^{\mathrm{PS}}\left(\boldsymbol{w}_n^{\mathrm{JS1}}, \boldsymbol{x}_k\right)=\frac{p\left(\boldsymbol{w}_n^{\mathrm{JS1}}, \boldsymbol{x}_k\right)}{\sum_{i=1}^{N} p\left(\boldsymbol{w}_i^{\mathrm{JS1}}, \boldsymbol{x}_k\right)}=\frac{p\left(\boldsymbol{w}_n^{\mathrm{JS1}}, \boldsymbol{x}_k\right)}{V} . \quad (18)$$

The definition of the constant $V$ is obvious from (18), and we have $\sum_{\mathrm{i}} p^{\mathrm{PS}}(\boldsymbol{w}_i^{\mathrm{JS1}}, \boldsymbol{x}_k) = 1$. Now, we can write (17) as

$$\mathrm{S}_{\mathrm{S}}\left(\boldsymbol{W}^{\mathrm{JS1}}\right)=\log(V)+\frac{1}{\boldsymbol{V}}\mathrm{S}^{\mathrm{SP}}\left(\boldsymbol{W}^{\mathrm{JS1}}\right) . \quad (19)$$

From (19), it can be seen that the minimization of $S_S(\boldsymbol{W}^{JS1})$ is equivalent to the minimization of $S^{SP}(\boldsymbol{W}^{JS1})$, so equation (14) can be formally used even in the cases when $N<K$, since $V$ represents the constant value that is controlled by the PSA part of the algorithm.

In the context of artificial neural network realization, usually we can use every sample only once – which means that we cannot use the successive optimization technique. The solution proposed here is based on a cooperative-competitive concept that is called the time oriented hierarchical method (TOHM). In this approach, two time scales are proposed (it can be easily modified for the case of multiple-time scales). The idea of multiple time scales is used in physics for simulation purposes [6, 32], and, also, in the superstatistics area [5].

In TOHM, on a faster time-scale, we calculate parameters of global interest (principal subspace), while on the slower time scale, we calculate parameters that are of local interest (principal eigenvectors). Concrete realization is based on the Lagrangian multiplier idea – a solution that should be obtained on the faster time scale is used as a constraint on the slower time-scale. This can be formally represented as the minimization of the following cost function

$$\mathrm{JS}(\boldsymbol{W}) = \mathrm{JS1}(\boldsymbol{W}) + \mu\,\mathrm{JS2}(\boldsymbol{W}), \qquad (20)$$

where $\mu$ is a real number, which is by absolute value smaller than one (since the JS2 should be realized on a slower time-scale). Here, the Lagrangian multiplier does not multiply the constraint part, but the objective one. It does not affect the solution since it represents the weight ratio of the different parts of the learning rule.

Several algorithms that perform PCA, and that are based on the TOHM method, are proposed in literature [16], [17], [19] and [20]. Without going into the details, in all those approaches the PCA algorithms are obtained from some known PSA algorithms together with the minimization of the Tsallis entropy, $S_T(q \approx 1)$. The value of $\mu$ was selected in all algorithms independently, after the stability analysis for every particular choice of JS1 and JS2.

We will also mention here that with the probabilistic definition given here, PCA can be used in a semi-supervised context, and can be successfully formulated for implementation on parallel computers. This is currently under investigation.

## VI. Experimental Results

In this section, we will present some results that illustrate the usefulness of the proposed concept. We are going to present some small scale experimental results that illustrate how the choice of the divergence/distance function influences the convergence speed and preciseness of the algorithm. Then, we will show some experimental results for scale robust PSA algorithm, based on BACH divergence. In the last part of this section, we are going to propose the BACH algorithm that represents modification of the single principal component analyzer, proposed by Oja [29].

### *A. Small scale simulation results*

Now, we will present some small scale simulation results. Here, we will analyze the behavior of several PSA learning algorithms implemented in the feedforward, linear, single layer neural networks with four inputs ($K$=4). We will consider two cases: first, we will consider the network with four outputs and after that we will consider the network with two outputs ($N$={2,4}). We will assume that input samples are representatives of some set $\Omega = \{\boldsymbol{x}_1, \boldsymbol{x}_2, \ldots\}$, and probability distributions $P$ and $Q$ are defined on $\Omega$, as $P = \{p(\boldsymbol{x}) \mid \boldsymbol{x} \in \Omega\}$ and $Q = \{q(\boldsymbol{x}) \mid \boldsymbol{x} \in \Omega\}$. Let's assume that $p^*$ and $q^*$ are defined as:

$$p^*(\boldsymbol{x}_k) = \boldsymbol{x}_k^{\mathrm{T}} \boldsymbol{x}_k \propto p(\boldsymbol{x}_k) \text{ and } q^*(\boldsymbol{x}_k) = \boldsymbol{y}_k^{\mathrm{T}} \boldsymbol{y}_k \propto p(\boldsymbol{x}_k) p(\boldsymbol{W} \mid \boldsymbol{x}_k),$$

where $\boldsymbol{y}_k=\boldsymbol{y}(k)$ is defined by equation (5).

We will consider several divergence/distance functions that were given in the Table I, and respective learning algorithms obtained by the gradient descent method presented in Table II.

Table II

| Divergence/ Distance name (Algorithm abbreviation) | Learning algorithm |
|---|---|
| Variational distance (A1) | $\Delta \boldsymbol{W} \sim \left(\boldsymbol{xy}^{\mathrm{T}} - \left(1-\delta(K,N)\right)\boldsymbol{W}\ \mathrm{diag}\left(\mathrm{diag}\left(\boldsymbol{yy}^{\mathrm{T}}\right)\right)\right)\mathrm{sign}\left(q^{*}-p^{*}\right)$ |
| Quadratic variational divergence (A2) | $\Delta \boldsymbol{W} \sim \left(\boldsymbol{xy}^{\mathrm{T}} - \left(1-\delta(K,N)\right)\boldsymbol{W}\ \mathrm{diag}\left(\mathrm{diag}\left(\boldsymbol{yy}^{\mathrm{T}}\right)\right)\right)\left(q^{*}-p^{*}\right)$ |
| Jeffrey's J-divergence (A3) | $\Delta \boldsymbol{W} \sim \left(\boldsymbol{xy}^{\mathrm{T}} - \left(1-\delta(K,N)\right)\boldsymbol{W}\ \mathrm{diag}\left(\mathrm{diag}\left(\boldsymbol{yy}^{\mathrm{T}}\right)\right)\right)\left(\log\left(\frac{q^{*}}{p^{*}}\right)+\frac{q^{*}-p^{*}}{p^{*}}\right)$ |
| Jensen-Shannon divergence (A4) | $\Delta \boldsymbol{W} \sim \left(\boldsymbol{xy}^{\mathrm{T}} - \left(1-\delta(K,N)\right)\boldsymbol{W}\ \mathrm{diag}\left(\mathrm{diag}\left(\boldsymbol{yy}^{\mathrm{T}}\right)\right)\right)\log\left(\frac{q^{*}+p^{*}}{2p^{*}}\right)$ |
| Hellinger distance (A5) | $\Delta \boldsymbol{W} \sim \left(\boldsymbol{xy}^{\mathrm{T}} - \left(1-\delta(K,N)\right)\boldsymbol{W}\ \mathrm{diag}\left(\mathrm{diag}\left(\boldsymbol{yy}^{\mathrm{T}}\right)\right)\right)\left(\frac{\sqrt{q^{*}}-\sqrt{p^{*}}}{\sqrt{p^{*}}}\right)$ |

In Table II, $\delta(K, N)$ represent Kronecker $\delta$ and $\Delta\boldsymbol{W}$ represents learning increment of the synaptic weight matrix $\boldsymbol{W}$. Learning methods are obtained by application of gradient descent on a particular divergence/distance measure. We can see that algorithms A1-A5 require only two global calculation circuits for calculation of the input and output energy ($p^{*}$ and $q^{*}$).

a) *Case N=4*

In the first set of experiments, we will adopt $N$=4. That means that the number of inputs and outputs is the same. This case is useful, for instance, when the PCA algorithm is used in the prewhitening step of some ICA algorithm (see e.g. [9]), or for matrix orthonormalization. The input signal in the *i*-th iteration (***x***(*i*)) is defined by the following equation (we use standard MATLAB notation)

$$a = \mathrm{diag}([1\ 0.8\ 0.6\ 0.4])\mathrm{randn}(4,\ 25000); \quad x(i) = 0.5(0.5 + 0.5\,\mathrm{rand}(1))\frac{a(i)}{\mathrm{norm}(a(i))} / pf;$$

and in that way the energy of the input signal is kept in the range (1:4, 1:1) if *pf*=1. This range is chosen arbitrarily. Limitation of the input energy is quite natural in technical and biological systems, and here, has the additional purpose to limit learning increments, since several of adopted learning rates depend inversely on the output signal power. A change of the value of *pf* influences the energy level. In all experiments the initial value of the synaptic matrix ***W*** was chosen as

$$\boldsymbol{W} = -0.05 + 0.2\ \mathrm{rand}(4,4).$$

The learning rates $\gamma$ were constant and chosen to achieve almost maximal convergence speed (learning rates were chosen experimentally – they were gradually increased until the learning algorithms became unstable, and then, the learning rate was reduced to 85% of the value that caused instability). After 12000 iterations, the learning rate was decreased 16 times. In the case of algorithm A1, γ=*lf**8*0.6, in the case A2 γ= *lf**8*2.7, for A3 γ= *lf**8*0.03, for A4 γ = *lf**8*0.8 and for A5 γ = *lf**8*0.96. In the following figures, the behavior of the algorithms is depicted for several values of the parameter *pf*. The *pf* was selected as 1, 2, 3, 4 and 5 respectively (*lf*=1). Figs. 4-8 represent the behavior of algorithms, A1-A5. Fig. 9 represents the behavior of the Subspace Learning Algorithm (SLA) [30]. Implementation of the SLA algorithm requires $K$ global calculation circuits (where $K$ is the number of inputs). The error was defined as

$$\text{Error} = \log\left(\text{abs}\left(\boldsymbol{I} - \boldsymbol{W}^{\mathrm{T}}\boldsymbol{W}\right)\right),$$

where $\boldsymbol{I}$ is $N$-dimensional identity matrix.

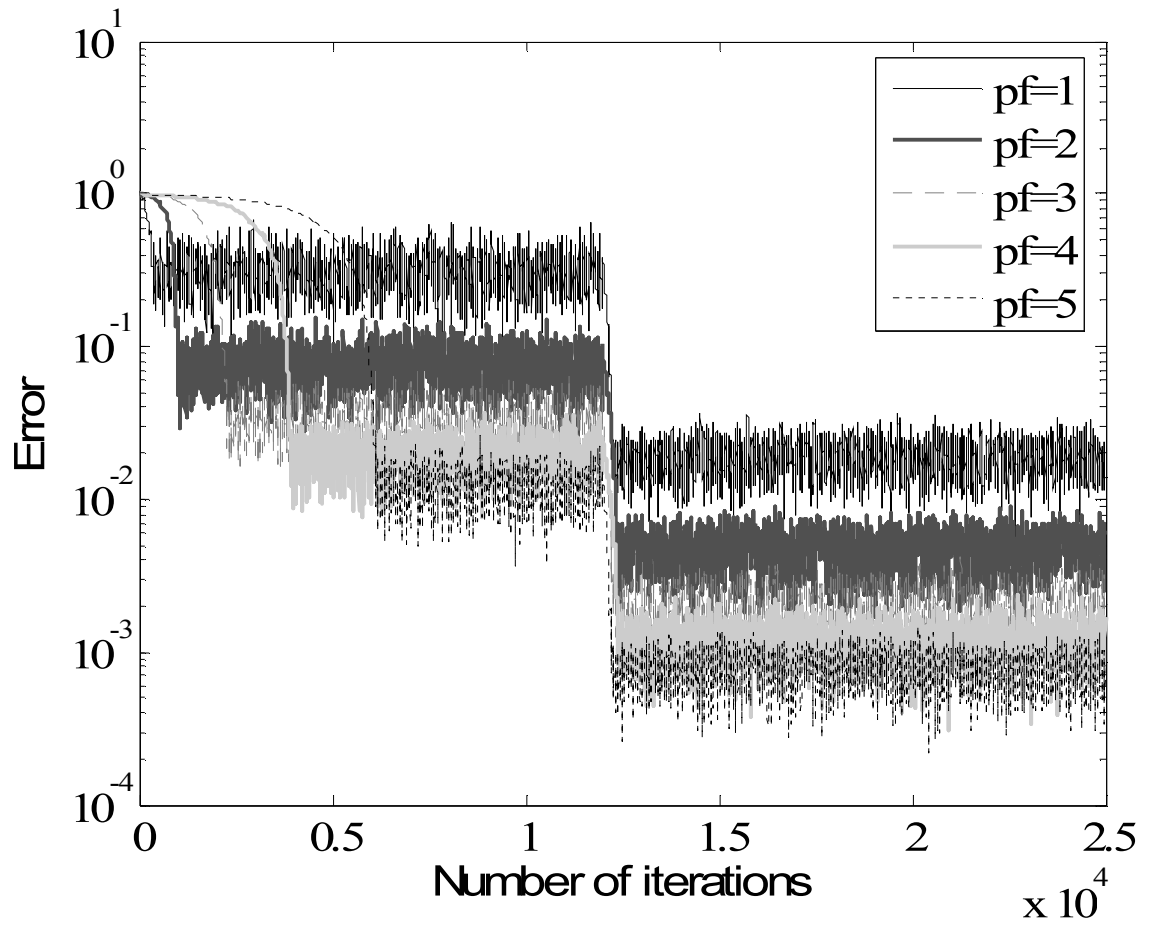

Fig. 4 Experimental results for (A1)

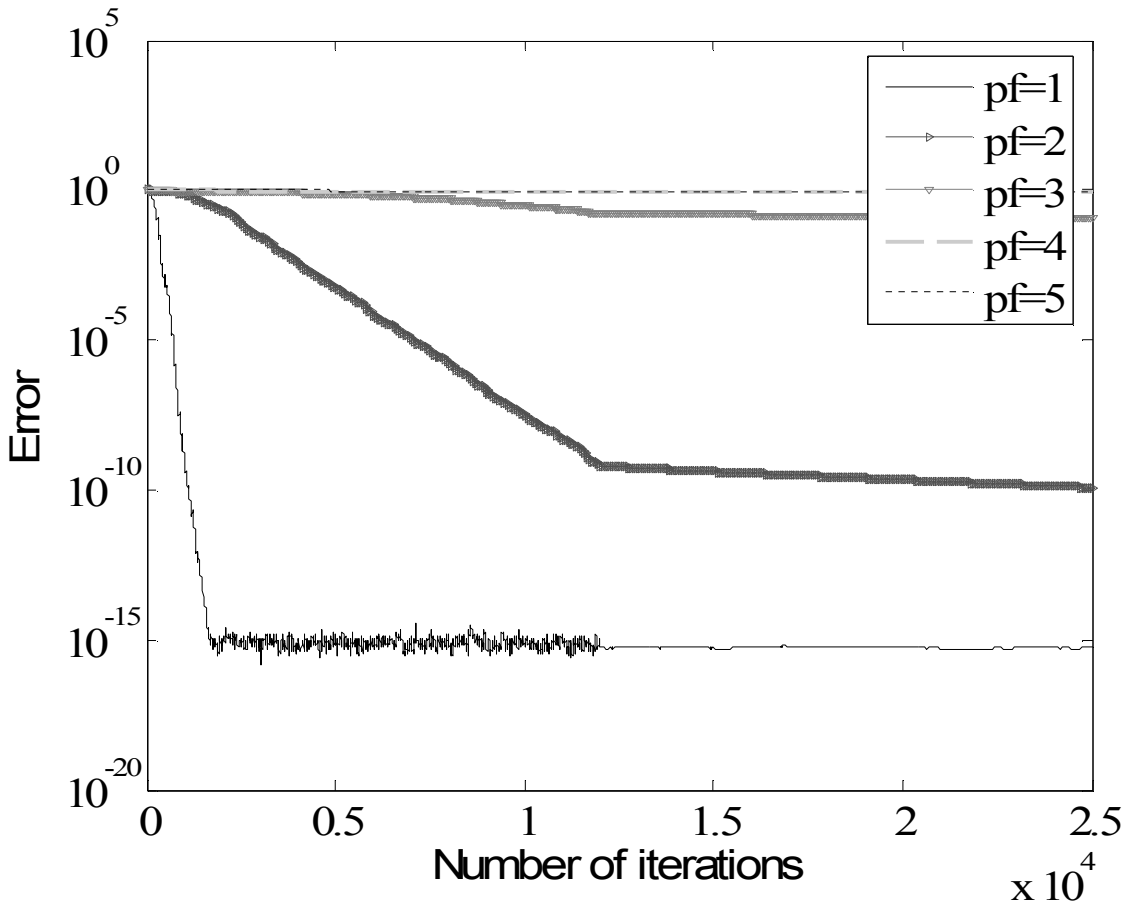

Fig. 5 Experimental results for (A2)

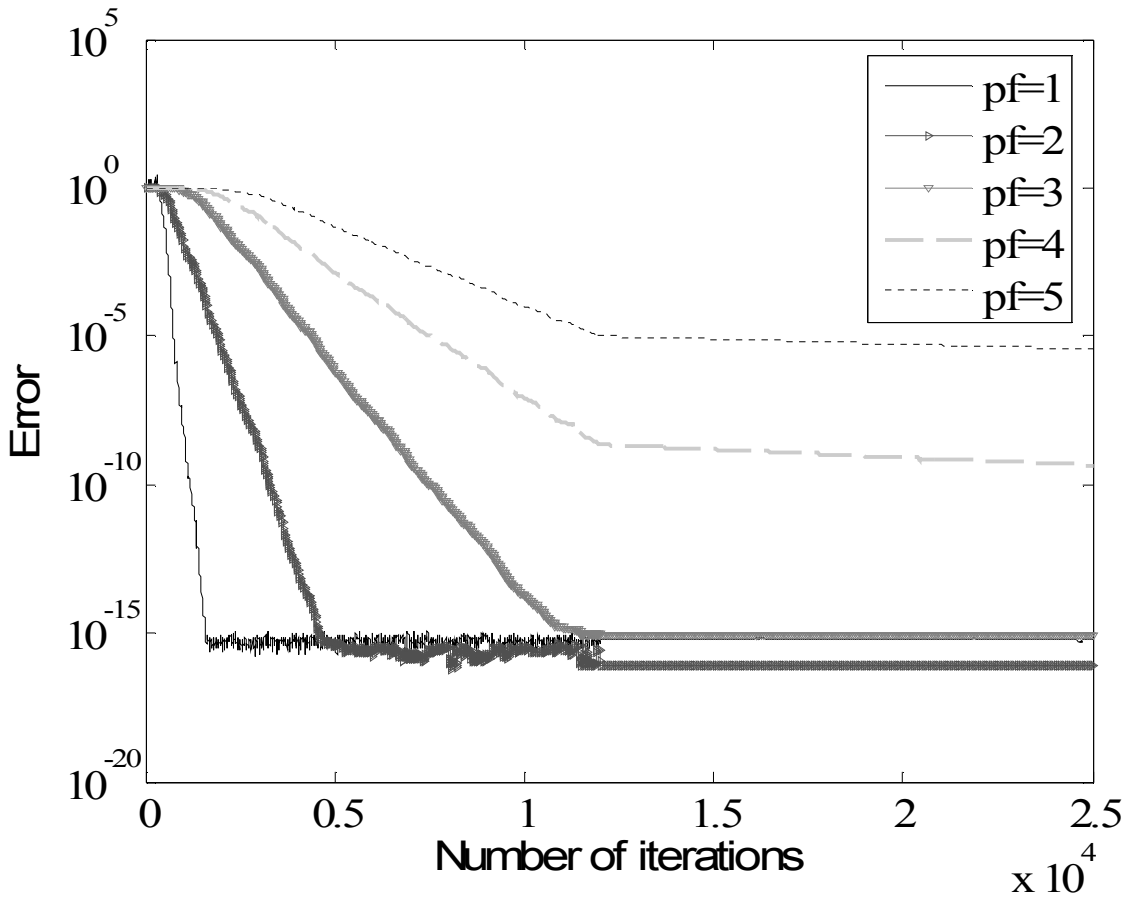

Fig. 6 Experimental results for (A3)

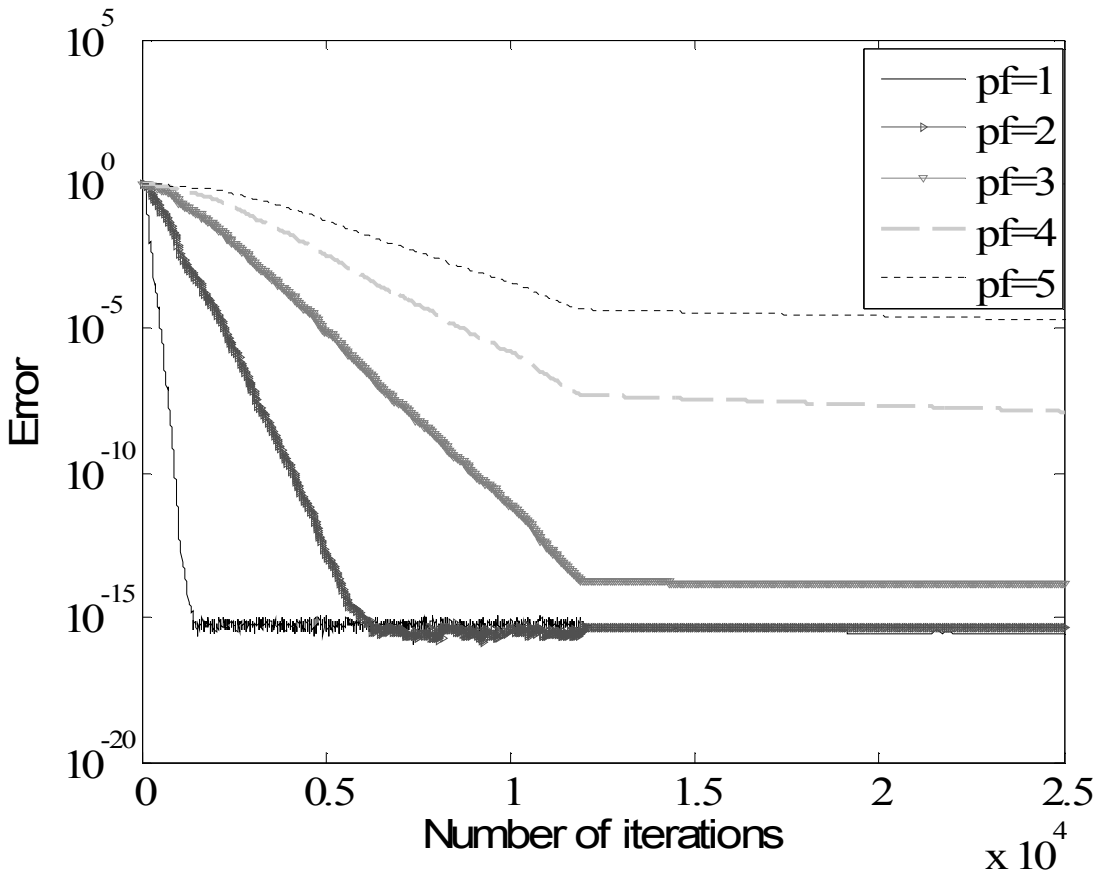

Fig. 7 Experimental results for (A4)

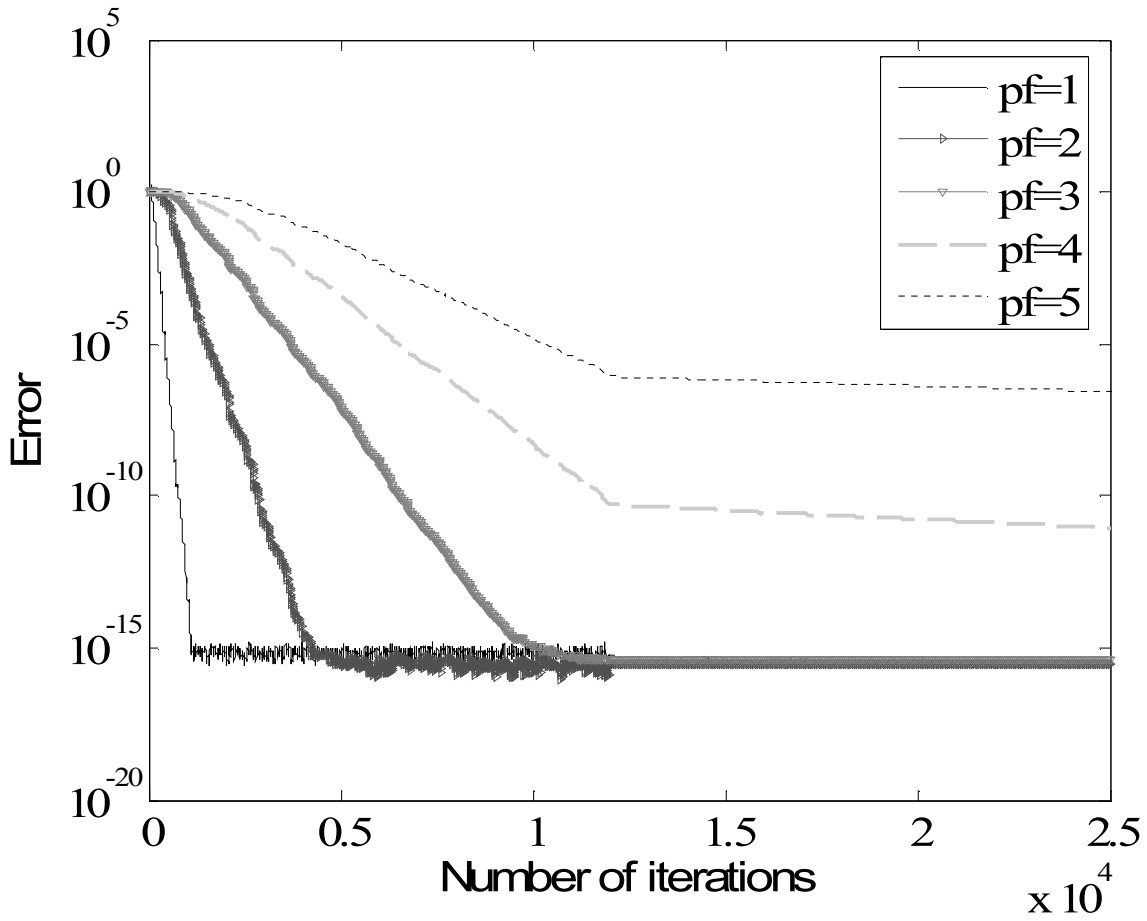

Fig. 8 Experimental results for (A5)

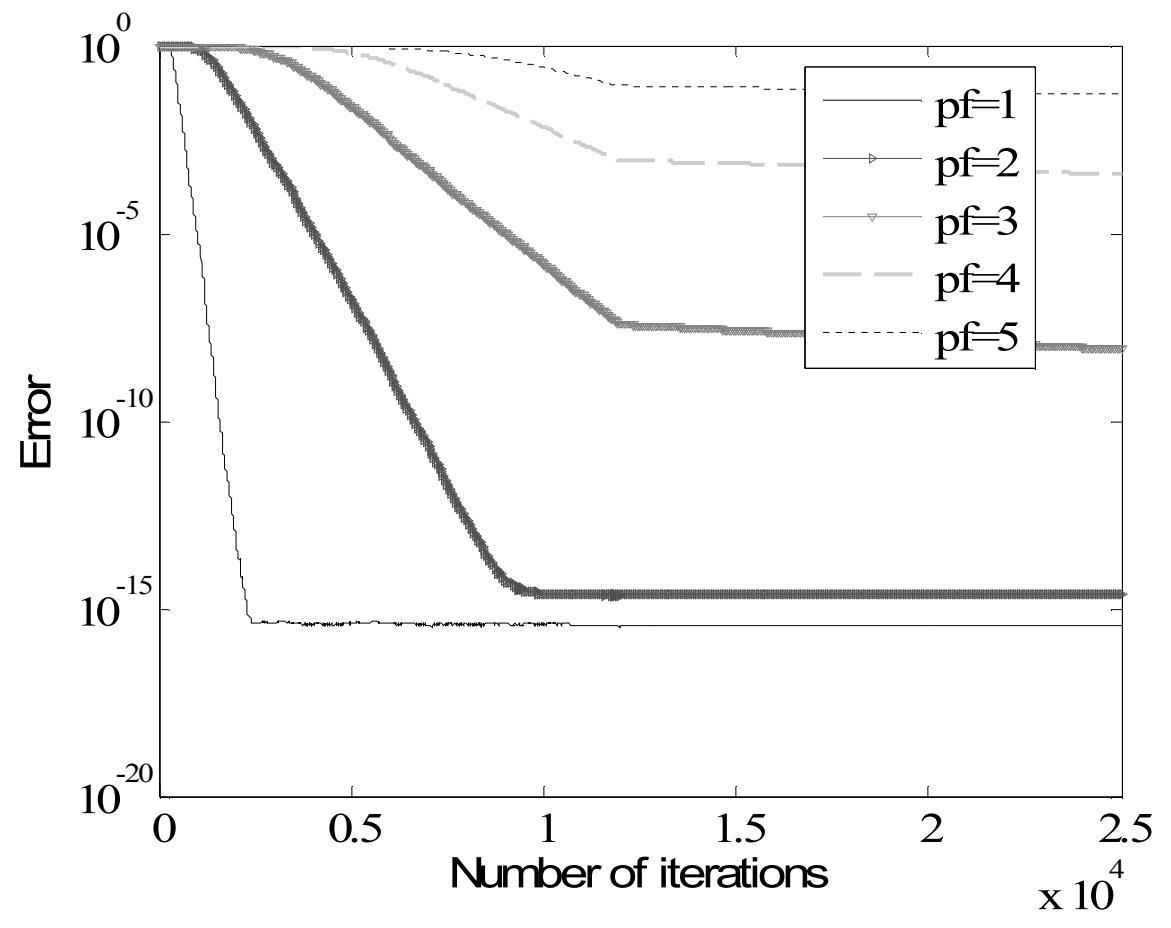

Fig. 9 Experimental results for (SLA)

From the experiments, we can conclude that the SLA algorithm has smoother behavior and better preciseness than all other presented algorithms. However, we can notice that this algorithm is sensitive to the level of input energy, since the convergence speed and preciseness of the algorithm are affected significantly with the change of the input energy level. It seems that algorithms A3-A5 are more robust in that sense. Algorithm A1 is interesting, since its behavior is different from all other algorithms – preciseness is improved with the decrease of the input energy level, while convergence speed is not severely reduced. In the papers that follow, we will show how we can modify algorithm A1 to obtain scaling robust learning algorithm, and achieve good preciseness. It also can be noticed that reduction of the learning rate, after some time, improves the preciseness and stability for all algorithms.

b) *Case N=2*

In the following set of experiments, we will adopt $N$=2. In this case the input signal is compressed. The input signal is defined as in the previous set. Initial learning rates for all algorithms are decreased 8 times. Variable parameter for experiments is learning rate that is held at the ratios 1:1/2:1/4:1/8:1/16 (*lf* =1, *lf*=1/2, *lf*=1/4, *lf*=1/8, *lf*=1/16 respectively, and *pf* =1). Again, after 12000 iterations learning rates were decreased 16 times. Figures 10, 11, 12, 13 and 14 depict behavior of the algorithms A1-A5, while Fig. 15 represents behavior of SLA algorithm.

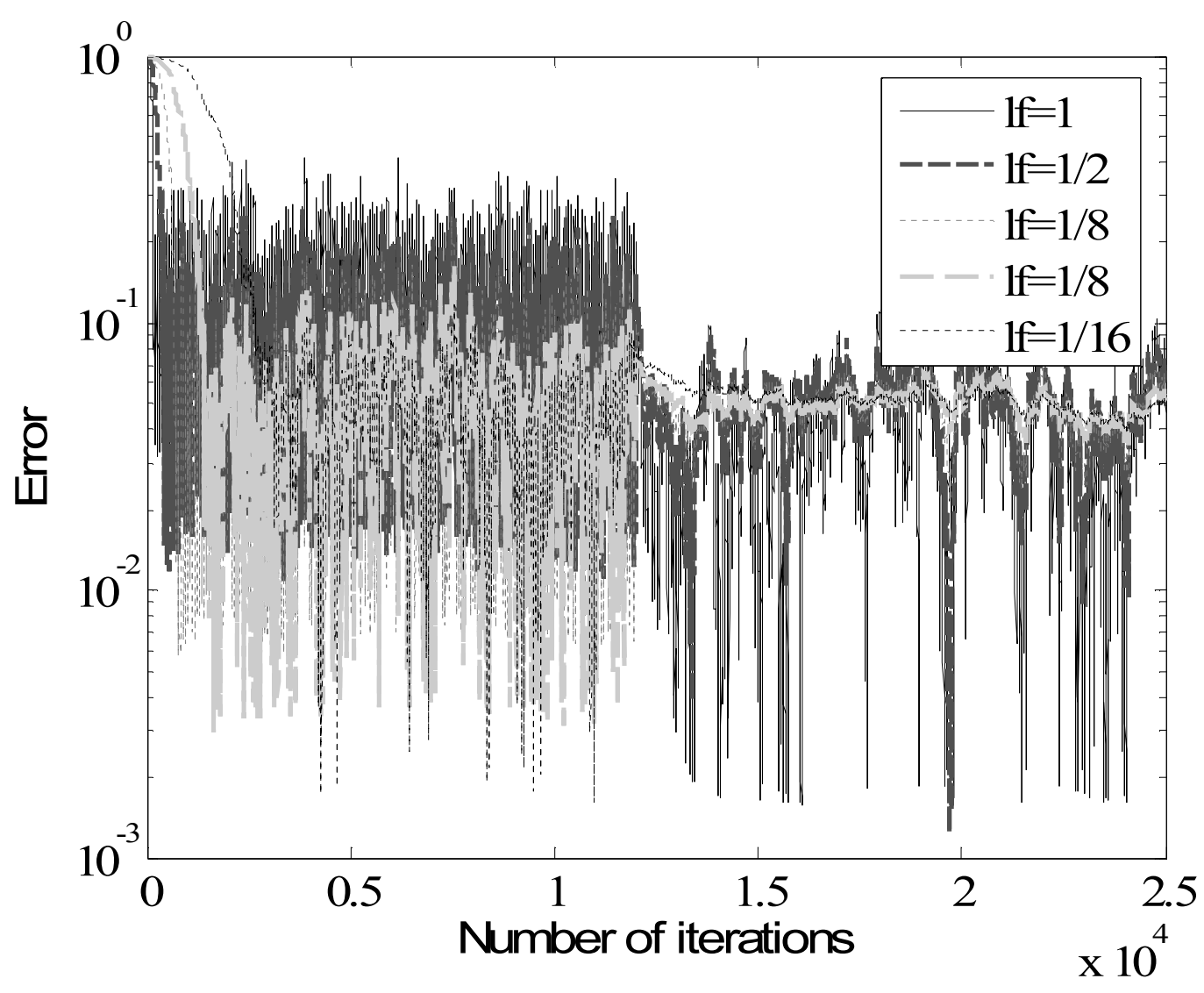

Fig. 10 Experimental results for (A1)

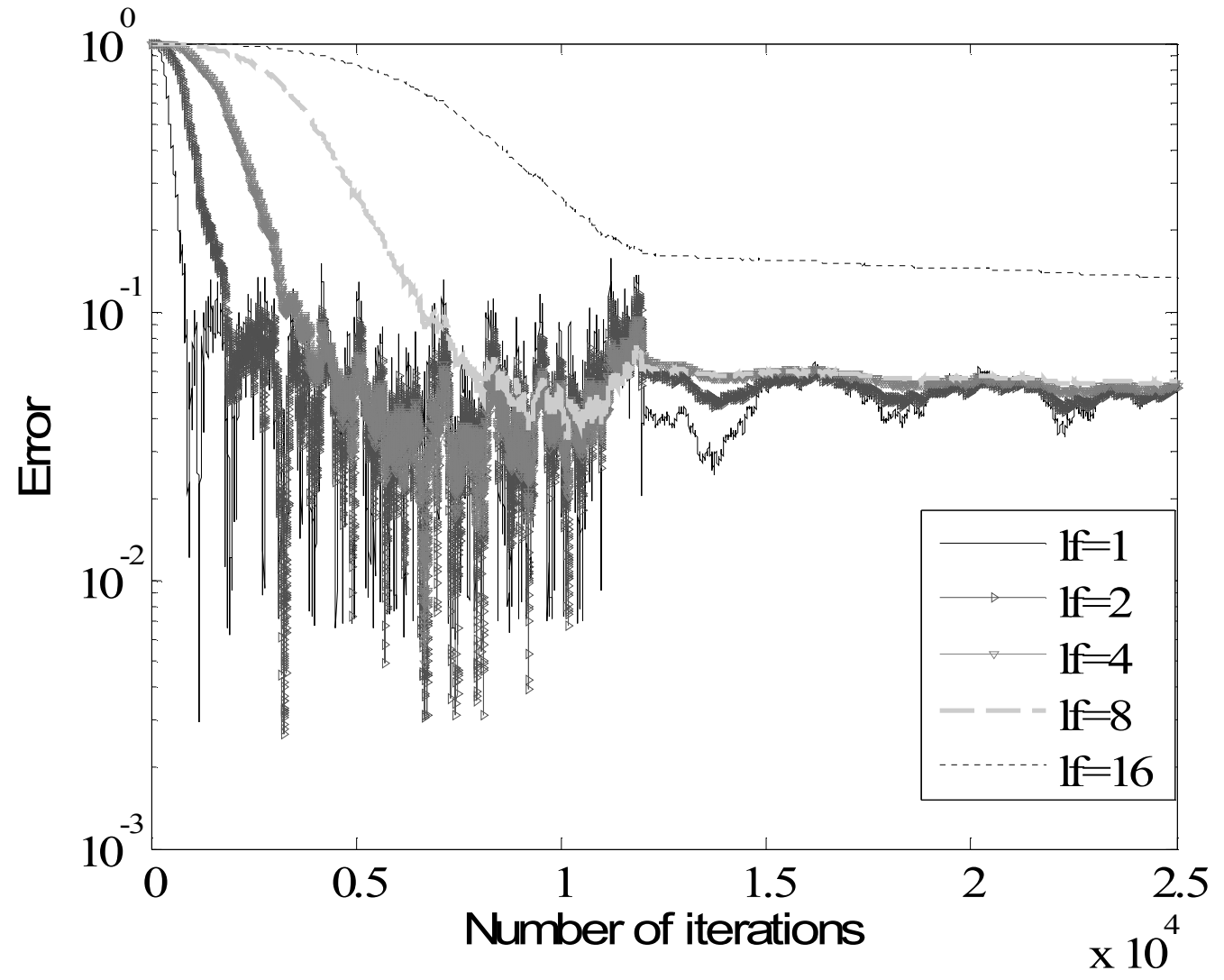

Fig. 11 Experimental results for (A2)

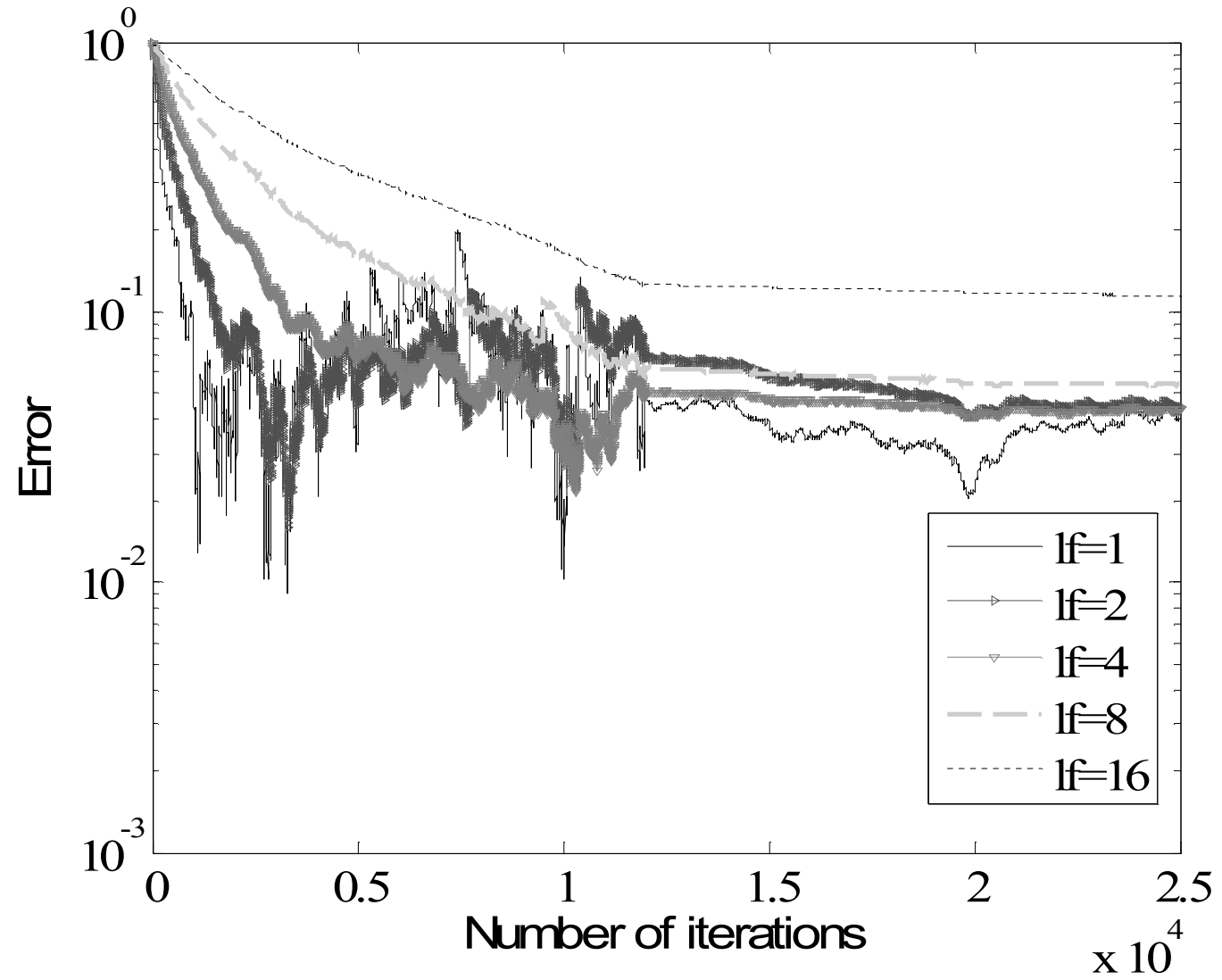

Fig. 12 Experimental results for (A3)

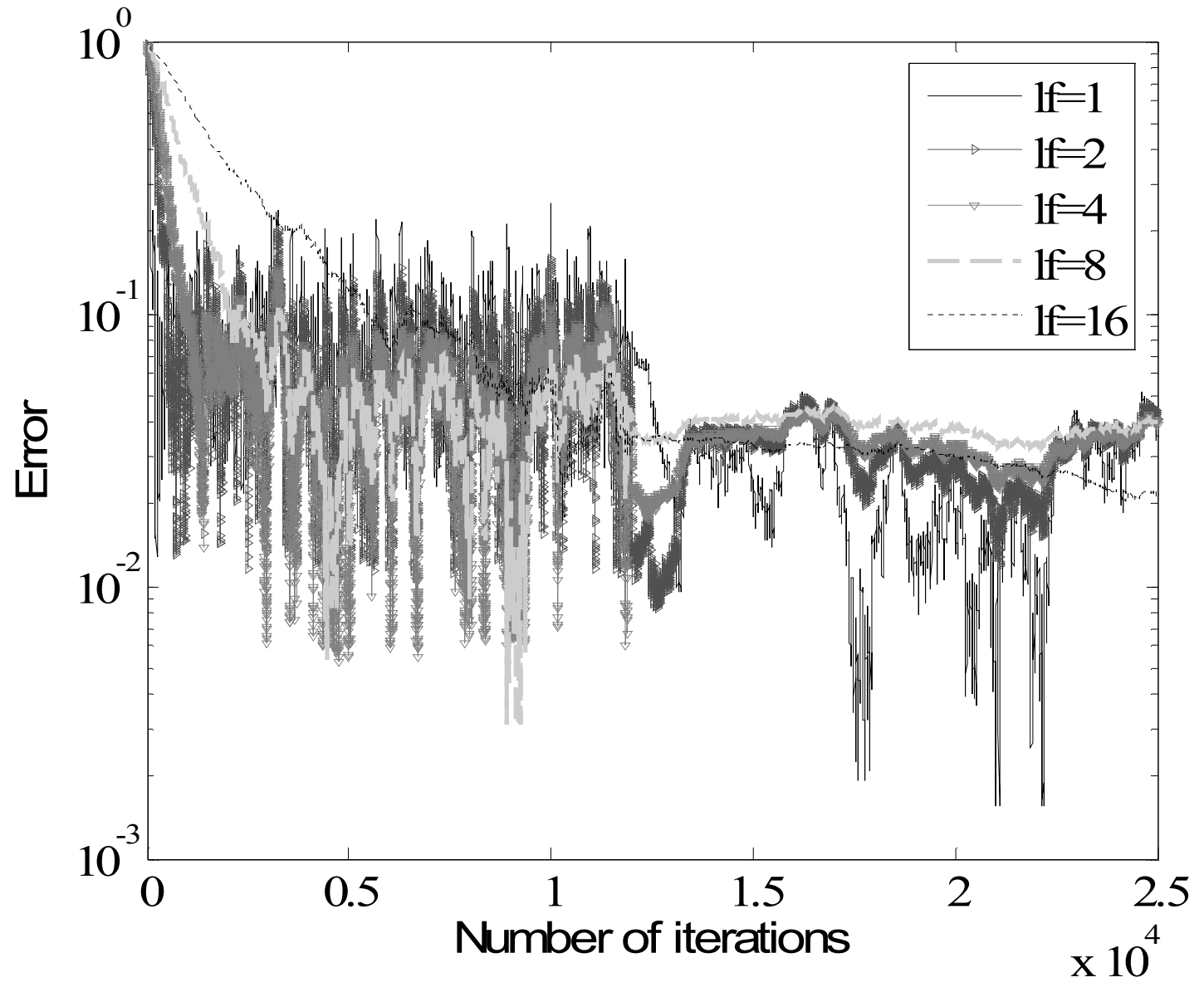

Fig. 13 Experimental results for (A4)

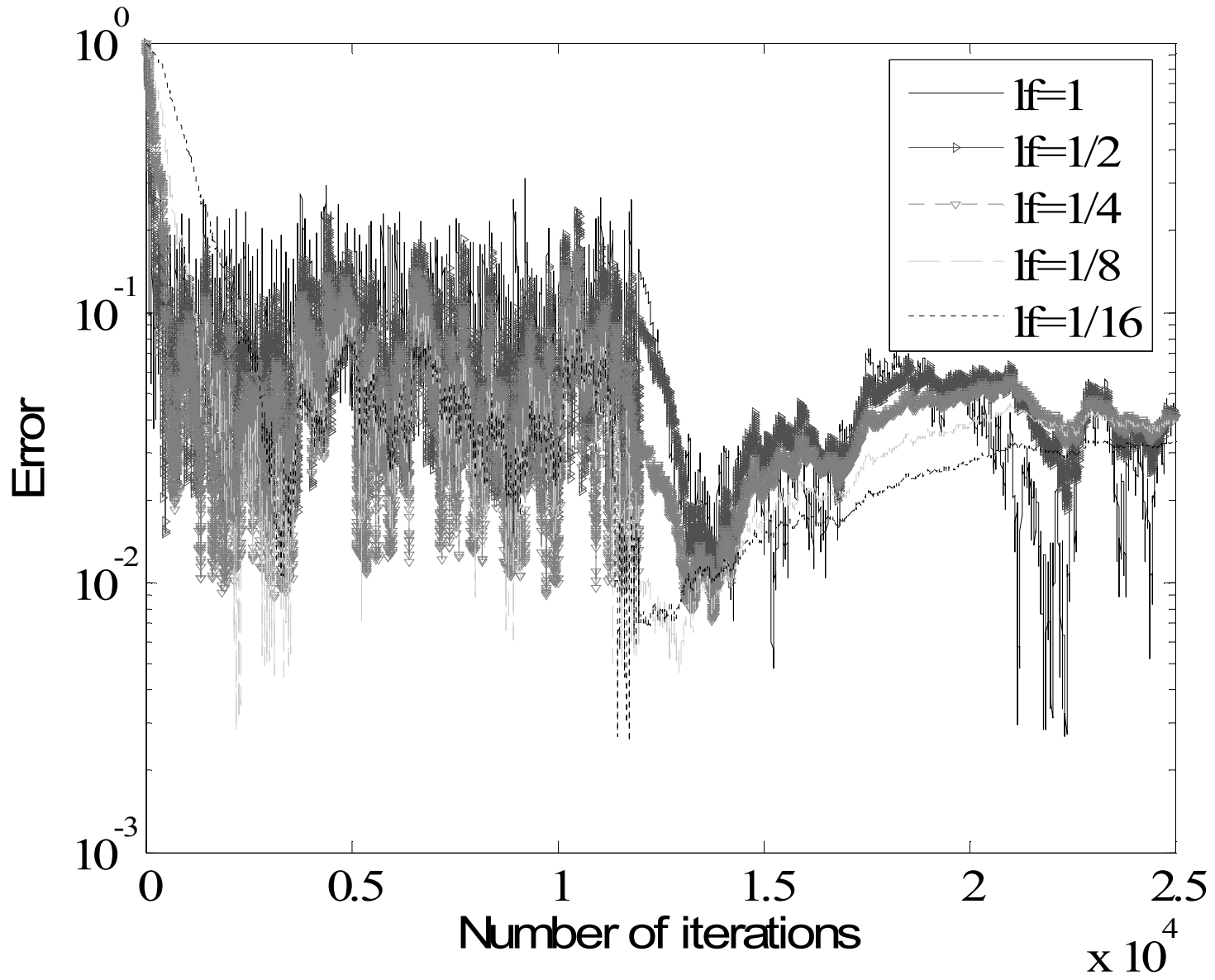

Fig. 14 Experimental results for (A5)

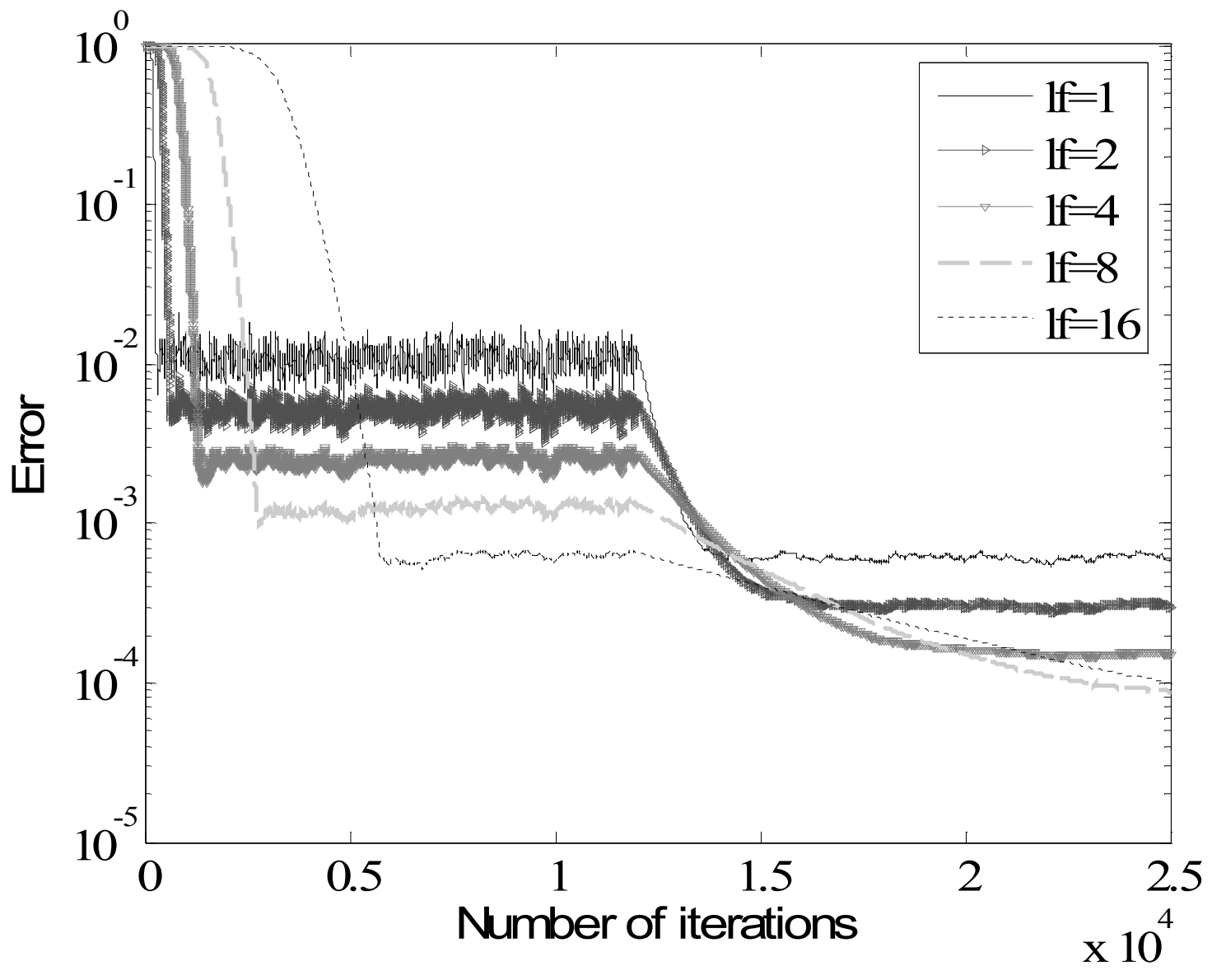

Fig. 15 Experimental results for (SLA)

From this set of experiments, we can conclude, again, that the SLA algorithm is the most precise for the same convergence speed, but its convergence speed is significantly affected by the decrease of the learning rate. We can see that the convergence speed of algorithms A1, A2 and A5 is much less affected by the change of the learning rate, comparing to the SLA algorithm. However, they are less precise. On the other side, algorithms A1-A5 are much more suitable for realization in parallel hardware, since they require only 2 global calculation circuits, irrespectively of the number of output neurons.

It seems that with careful choice of the divergence/distance function we can achieve very similar convergence speed and preciseness as in the case of the SLA algorithm, and at the same time, still, have local implementation of the learning rule.

*B. BACH divergence based scale robust PSA learning algorithm*

Now, we are going to introduce the BACH divergence as

$$D_{\mathrm{BACH}}(P \parallel Q) = \mathrm{E}\left(\left(p^b - q^b\right)^2\right),$$

where *b* is a positive real number (in this section we are going to assume that *b* is smaller than 1). We can see that the Hellinger distance represents special case where *b*=0.5. Here, we assume *b*=0.025. Then, analyzing neural network in the set-up that is explained in subsection *A.a*), we will obtain the result depicted in Fig. 16 (after implementation of gradient descent).

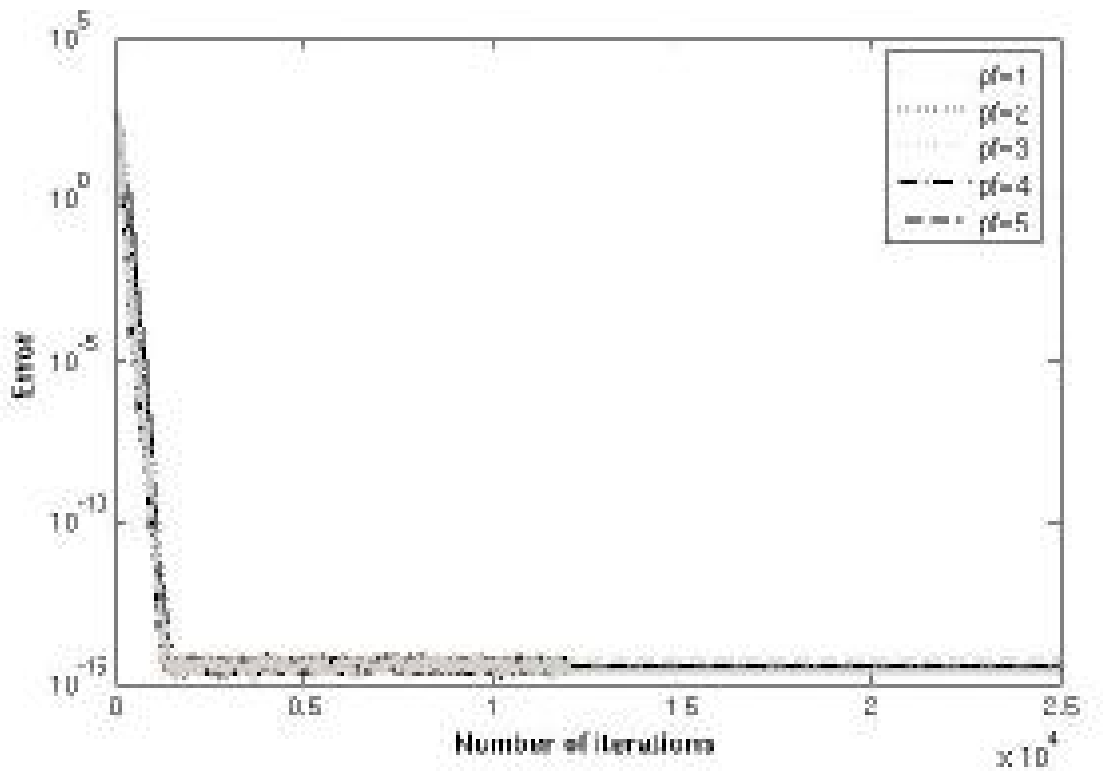

Fig. 16. Experimental results for BACH divergence (4-dimensional case without compression)

We can clearly see that speed and preciseness of the algorithm are almost not affected at all, despite the change of the energy in the input signal in the ratio 1 to 1/25. This could be seen as a good characteristic of the proposed algorithm, comparing it, for instance, to SLA algorithm.

In the two following pictures, we can see the behavior of the algorithm in the cases when the input signal dimension is 100 (Fig. 17) and 1000 (Fig. 18), while the output dimension is 4. The proposed algorithm, despite low memory requirements and small number of the global calculation circuits, is useful in the high dimensional input case.

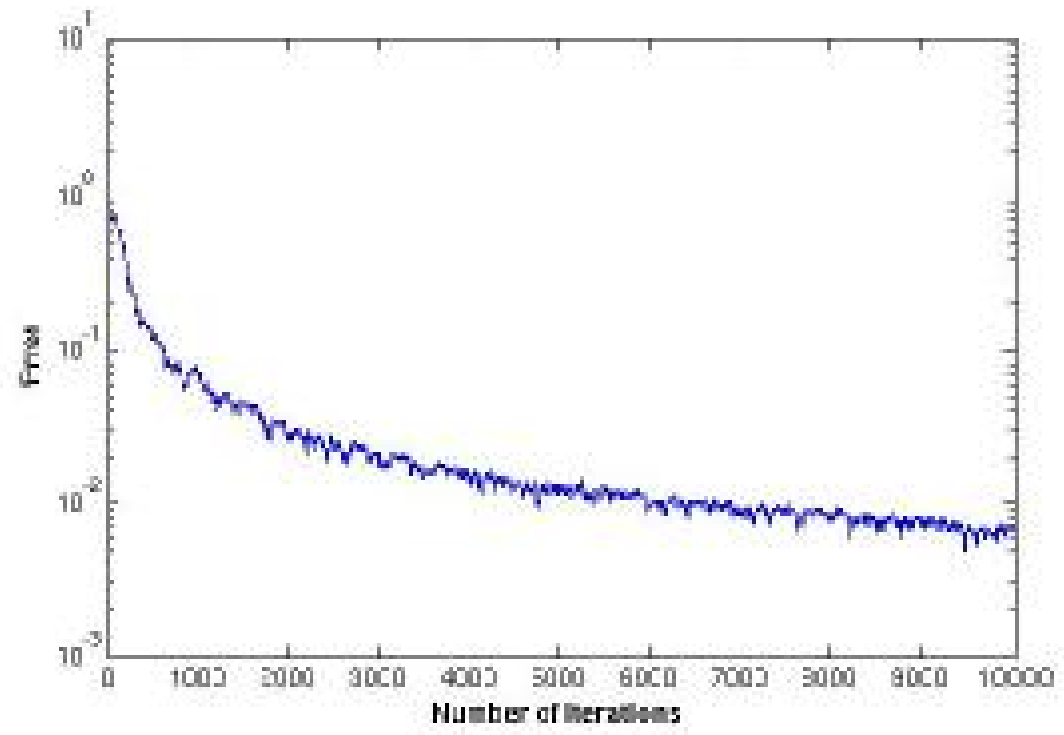

Fig. 17 Experimental results for the BACH divergence (100-dimensional input, 4-dimensional output)

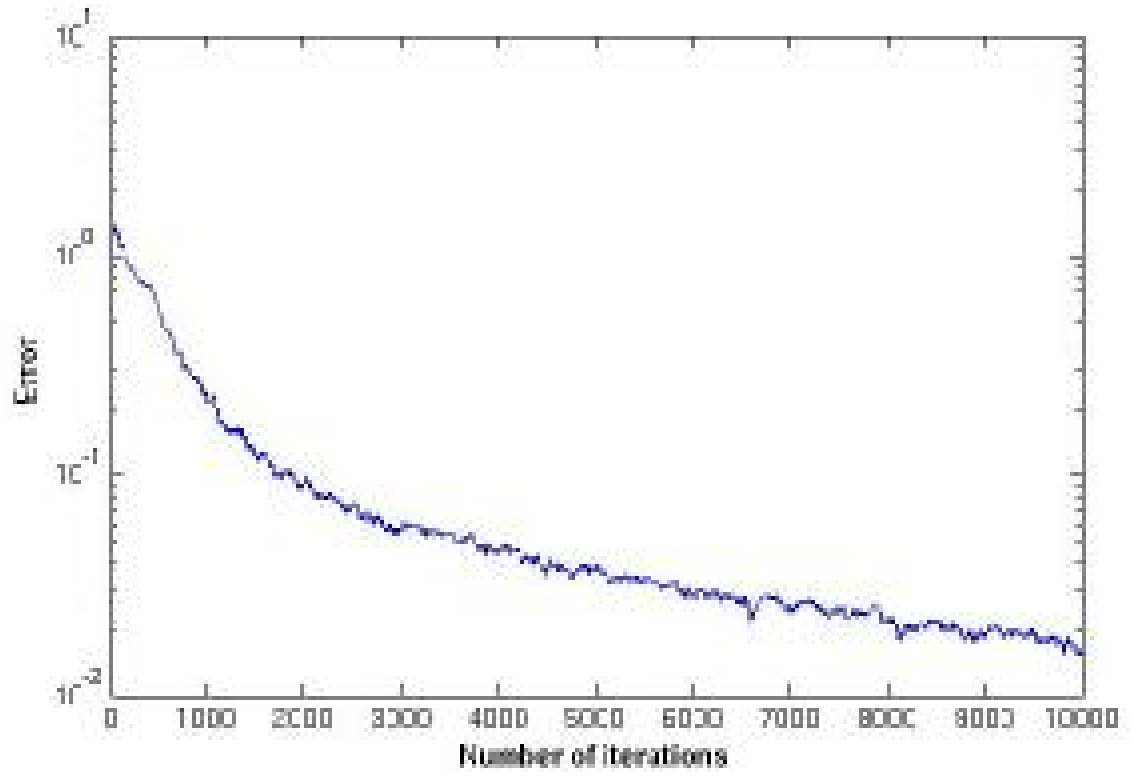

Fig. 18 Experimental results for the BACH divergence (1000-dimensional input, 4-dimensional output)

*C. BACH learning algorithm for single principal component extraction*

In this section, we will propose a new algorithm for extraction of the principal component. We will call it the BACH algorithm. It is based on minimization of the BACH divergence. For illustration, we will assume that we have 16-dimensional input signal $\boldsymbol{x}$ sampled from uniform distribution. Our learning rule is given as

$$\Delta\boldsymbol{w} \sim (\boldsymbol{x}y^{\mathrm{T}} - \boldsymbol{w}y^2)((\boldsymbol{x}^{\mathrm{T}}\boldsymbol{x})^b - (y^{\mathrm{T}}y)^b)/(y^{\mathrm{T}}y)^{1-b}$$

where $b$ is positive real number. Here, we select $b$=0.025. Without going into details, we can remove term $((\boldsymbol{x}^{\mathrm{T}}\boldsymbol{x})^b - (y^{\mathrm{T}}y)^b)$, so BACH learning rule can be represented as

$$\Delta\boldsymbol{w} \sim \frac{(\boldsymbol{x}y^{\mathrm{T}} - \boldsymbol{w}y^2)}{(y^{\mathrm{T}}y)^{1-b}},$$

We can see that it represents modification of the famous Oja's learning rule [29], which is, as we are going to see from the illustration example, robust to change of the input signal energy level, which significantly simplifies selection of the learning rate. In the Fig. 19, we can see behavior of the BACH (solid line) and Oja's algorithm for the "nominal" energy level.

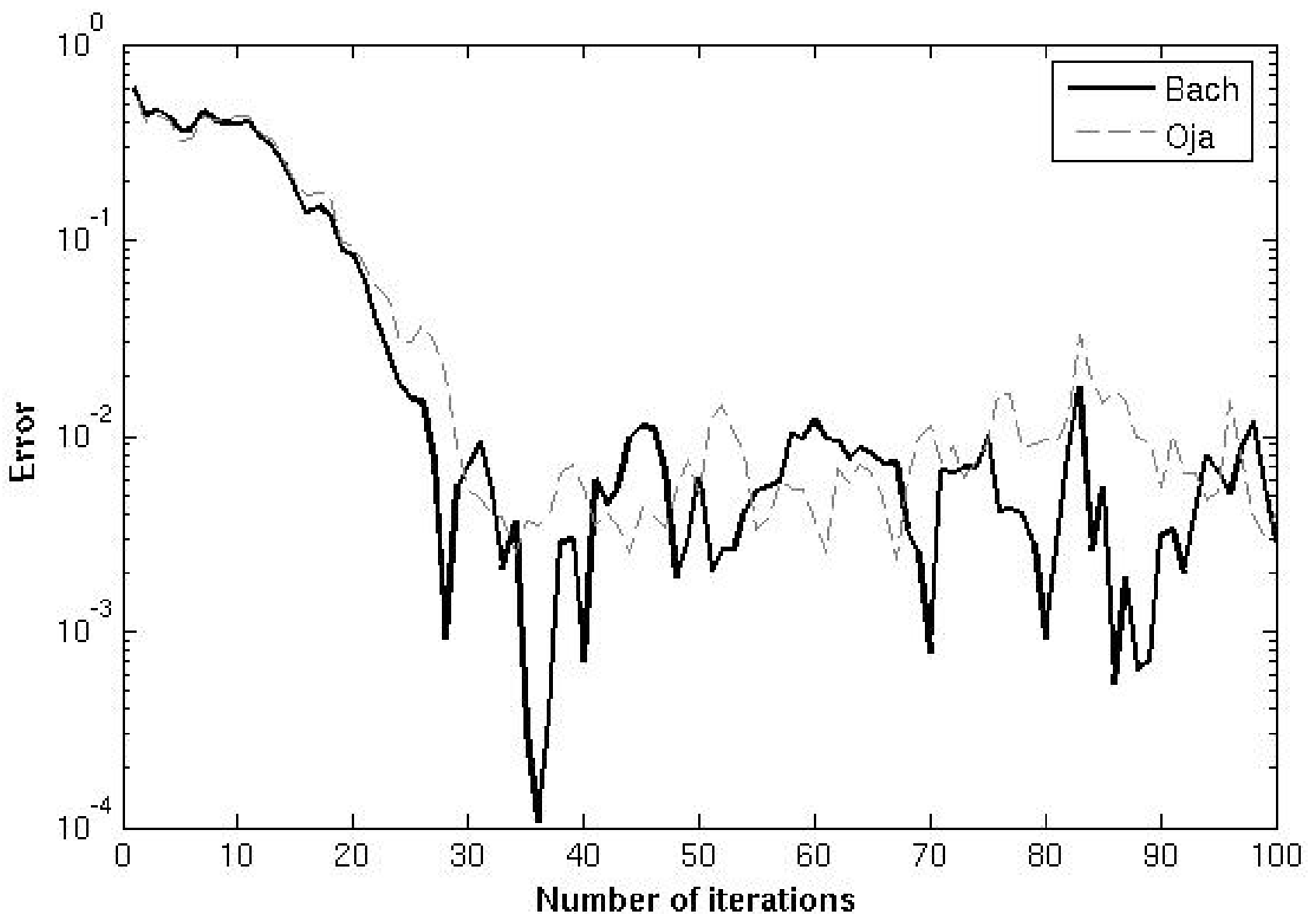

Fig. 19 Experimental results for BACH PCA algorithm (16-dimensional input, 4-dimensional output)

We can see that both algorithms behave similarly. However, if we decrease the input energy 100 times, and keeps learning rates unchanged, we can see (Fig. 20) that convergence speed of the BACH algorithm is only slightly affected by input signal energy change, while Oja's algorithm becomes very slow. Although it is not going to be presented here, it can be said that BACH algorithm will not be significantly affected, even in the case when the input energy is decreased by a several order of magnitudes (e.g. billion times).

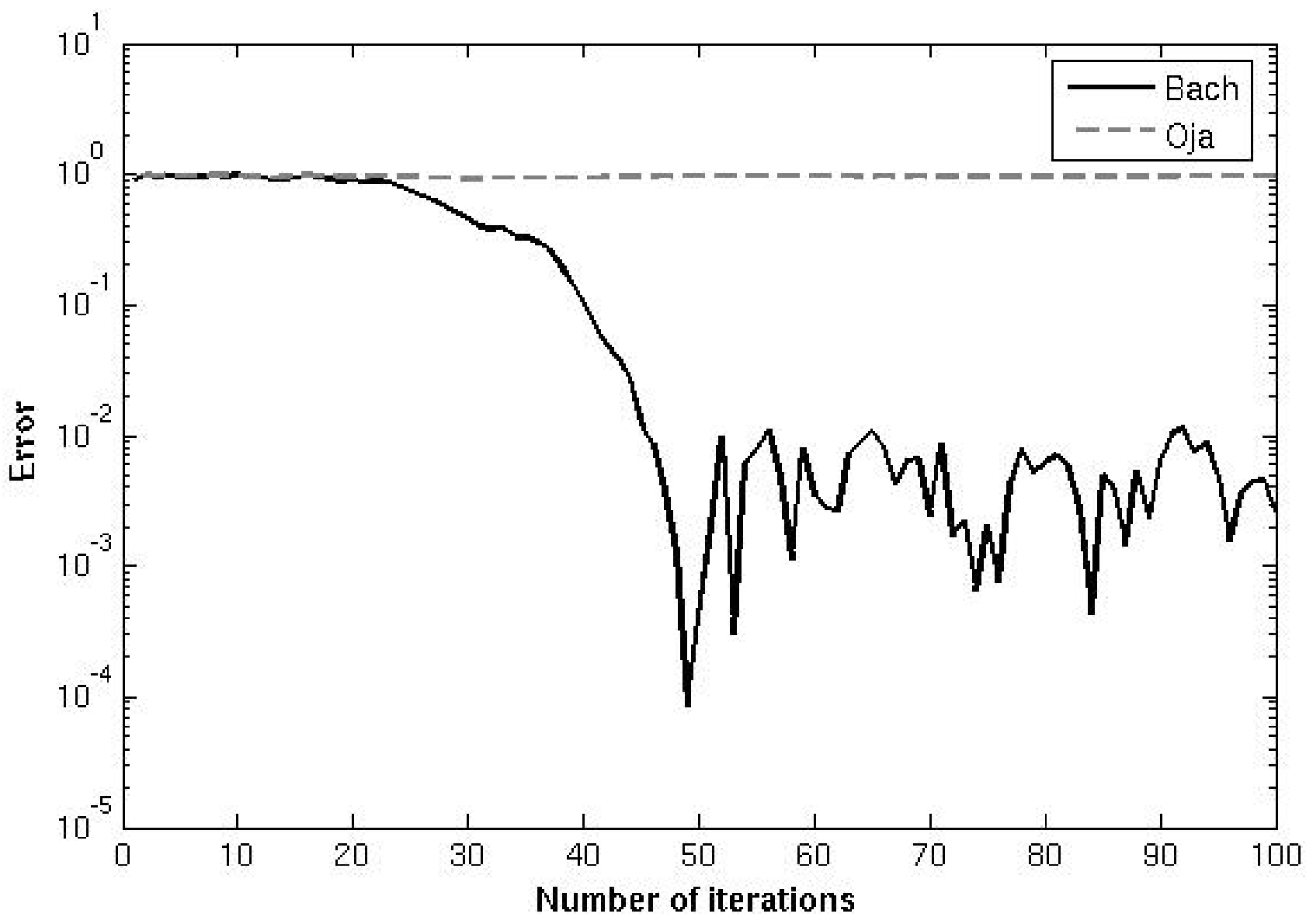

Fig. 20 Experimental results for BACH PCA algorithm (input signal energy decreased 100 times)

## VII. Conclusion

In this paper, we proposed a probabilistic principal subspace analysis (starting from MHO algorithm), as well as brief introduction of probabilistic principal component analysis, both based on the Born rule. Here we only considered derivation of symmetric algorithms. A new simple geometrical representation of the Born rule, named JoyStick Probability Selector was presented, too. Proposed probabilistic PCA can be solved off-line as a successive optimization problem. In on-line implementation, the problem is solved by the introduction of two distinct time-scales. By, selecting different divergence/distance functions, it is possible to create a big number of PSA algorithms. In that way, it is possible to create algorithms that could be potentially optimal from the point of view of convergence speed, preciseness, complexity of hardware implementation, locality of calculations etc. By selection of different PSA algorithms and different entropy functions, it is possible to create an even bigger number of algorithms that can be used for PCA, and that have some specific features. With this, probabilistic definition, PCA can be used in semi-supervised context and can be successfully implemented in parallel computation machines. This is currently under investigation.

From the presented experimental results, we can see that it is possible to create the “optimal” divergence/distance function, which can result in good algorithm characteristics (e.g. convergence speed, preciseness, scaling robustness) and still have a realization which has only two (or less) global calculation circuits. Like illustration, we presented input signal scale robust principal subspace algorithm, and BACH algorithm. A small number of global calculation circuits is a very useful characteristic of the proposed algorithms for successful implementation in parallel hardware, or for simple implementation on GPU platforms.

The proposed approach can be easily modified to give the definition of the probabilistic minor subspace analysis (MSA) and the minor component analysis (MCA).

It is possible to show that a similar way of reasoning can lead to a new probabilistic definition of the independent component analysis (ICA). By doing this, it is possible to give different interpretations of some existing ICA solutions and to propose new ones. One possible way to do it has been proposed recently in [33].

## REFERENCES

[1] J-H. Ahn, J-H. Oh and S. Choi. Learning principal directions: integrated-square-error minimization. *Neurocomputing*, 70: 1372-1381, 2007.

[2] S.M. Ali and S.D. Silvey. A general class of coefficients of divergence of the distribution from another. *J. Roy. Statistics. Soc.* Ser B, 28: 131-142, 1966.

[3] C. Archambeau, N. Delanney and M. Verleysen. Robust probabilistic projection. In: Proc. 23rd *International Conference on Machine Learning*, Pittsburg, USA, 2006.

[4] P. Baldi and K. Hornik. Learning in linear neural networks: A survey. *IEEE Trans. Neural Networks*, 6: 837-858, 1995.

[5] C. Beck. Superstatistics: theory and applications. *Continuum Mech. Thermodyn.*, 16:293-304, 2004.

[6] C.M. Bishop. Pattern Recognition and Machine Learning, Springer, 2006.

[7] T. Chen, E. Martin and G. Montague. Robust probabilistic PCA with missing data and contribution analysis fro outlier detection. *Computational Statistics and Data Analysis*, 53: 3706-3716, 2009.

[8] T.-P. Chen and S.-I Amari. Unified stabilization approach to principal and minor components. *Neural Networks*, 14: 1377-1387, 2001.

[9] A. Cichocki and S.-I. Amari. Adaptive Blind Signal and Image Processing – Learning Algorithms and Applications. John Wiley and Sons, 2003.

[10]M. Collins, S. Dasgupta, and R.E. Schapire. A generalization of principal component analysis to the exponential family. *Advances in Neural Information Processing Systems*, 14:617-624, 2002.

[11]I. Csiszár. A note on Jensen’s inequality. Studia. Sci. Math. Hungar., 1:185-188, 1966.

[12] Y. Fang and M.K. Jeong. Robust probabilistic multivariate calibration model. *Technometrics*, 50: 305-316, 2008.

[13]P. Földiák. Adaptive network for optimal linear feature extraction. *IJCNN'89*, 401-405, 1989.

[14]A. M. Gleason. Measure on the closed subspaces of a Hilbert space. *J. Math. & Mechanics,* 6: 885-893, 1957.

[15]H. Hotteling. Analysis of a complex of statistical variables into principal components. *Journal of Educational Psychology*, 24: 417-441, 1933.

[16]M. Jankovic and H. Ogawa. Time-Orineted hierarachical method for computation of principal components using subspace learning algorithm, *Int. J. Neural Systems*, 14: 313-324, 2004.

[17]M. Jankovic and B. Reljin. Neural learning on Grassman/Stiefel principal/minor submanifold, EUROCON 2005, 249-252, 2005.

[18]M. Jankovic and H. Ogawa. Modulated Hebb-Oja learning rule – A method for principal subspace analysis. *IEEE Trans. on Neural Networks,* 17: 345-356, 2006.

[19]M. Jankovic and B. Reljin. General stochastic approximation and neural learning on principal Stiefel submanifold”, *WSEAS Transactions on Information Science and Applications*, 3: 1195-1201, 2006.

[20]M. Jankovic and M. Sugiyama. A multipurpose linear component analysis method based on modulated Hebb Oja learning rule. *IEEE Signal Processing Letters*, 15: 677-680, 2008.

[21]M. Jankovic and M. Sugiyama. Tensor based image segmentation. *Ninth Symposium on Neural Network Applications in Electrical Engineering* (*NEUREL 2008*), 67-73, 2008.

[22]M. Jankovic and M. Sugiyama. Probabilistic principal component analysis based on joystick probability selector. *International Joint Conference on Neural Networks* (*IJCNN 2009*), 1414-1421, 2009.

[23]I. T. Jolliffe. Principal Component Analysis. 2nd ed., Springer –Verlag, New York, 2002.

[24]R. V. Kadison. The Pythagorean Theorem: I. The finite case. Proceedings of the National Academy of Sciences the United States of America, vol. 99., no. 7, pp. 4178-4184, 2002.

[25]R. V. Kadison. The Pythagorean Theorem: II. The infinite discrete case. Proceedings of the National Academy of Sciences the United States of America, vol. 99., no. 8, pp. 5217-5222, 2002.

[26]J. Karhunen, E. Oja, L. Wang, R. Vigario and J. Joutensalo. A class of neural networks for independent component analysis, *IEEE Trans. on Neural Networks*, 8: 486-504, 1997.

[27]D.J.C. MacKay. Information Theory, Inference and Learning Algorithms. Cambridge University Press, 2003.

[28]M.A. Nielsen and I.L. Chuang. Quantum Computation and Quantum Information. Cambridge University Press, 2000.

[29] E. Oja. A simplified neuron model as a principal component analyzer. *J. Math. Biol.*, 15:267-273, 1982.

[30]E. Oja. Neural networks, principal components, and subspaces. *Int. J. Neural Syst.*, 1: 61–68, 1989.

[31]K. Pearson. On lines and planes of closest fit to systems of points in space. *Philosophical Magazine*, 2:559-572, 1901.

[32]J.C. Sexton and D.H. Weingarten. Hamiltonian evolution for the hybrid Monte Carlo algorithm, *Nuclear Physics B*, 380: 665-677, 1992.

[33] T. Suzuki and M. Sugiyama. Estimating squared-loss mutual information for independent component analysis. *Independent Component Analysis and Signal Separation*, *Lecture Notes in Computer Science*, 5441: 130-137, Berlin, Springer, 2009.

[34]M. Tipping and C.M. Bishop. Probabilistic principal component analysis. *Journal of the Royal Statistical Society, Series B*, 61, part 3: 611-622, 1999.

[35]M. Tipping and C.M. Bishop. Mixtures of probabilistic principal component analysers. *Neural Computation*, 11: 443-482, 1999.

[36]C. Tsallis. Possible generalization of Boltzmann-Gibbs statistics. *J. Statistical Physics,* 52: 479-478, 1988.

[37]M.K. Warmuth and D. Kuzmin. Bayesian generalized probability calculus for density matrices. *To appear in Machine Learning*, 2009.

[38]L. Wolf. Learning using the Born rule. Technical report *MIT-CSAIL- TR-2006-036*, 2006.

[39]W. K. Wootters. Statistical distance and Hilbert space. *Physical Review D.*, 23:357-362, 1981.